\documentclass{article}

\usepackage{iclr2027_conference,times}
\usepackage[utf8]{inputenc} 
\usepackage[T1]{fontenc}    
\usepackage{hyperref}       
\usepackage{url}            
\usepackage{booktabs}       
\usepackage{amsfonts}       
\usepackage{nicefrac}       
\usepackage{microtype}      
\usepackage{xcolor}         
\newcommand{\rev}[1]{{#1}}
\hypersetup{hidelinks,pdftitle={NeuroMotion: Aligning EEG to Discrete Motion Primitives for Continuous Motion Decoding},pdfauthor={Yi Liu, Qi Zhang, HaoLin Li, Lai Zhong Yuan, Lei Zhu, Heng Tao Shen}}
\usepackage{amsmath}
\usepackage{graphicx}
\usepackage{multirow}
\usepackage{enumitem}

\title{NeuroMotion: Aligning EEG to
Discrete \\Motion Primitives for
Continuous Motion \\Decoding}

\author{%
  \normalfont\small Yi Liu$^{1}$ \quad Qi Zhang$^{1,2}$ \quad HaoLin Li$^{1}$\\[2pt]
  Lai Zhong Yuan$^{3}$ \quad Lei Zhu$^{1,2}$ \quad Heng Tao Shen$^{1}$\\[4pt]
  {\footnotesize $^{1}$Tongji University \qquad $^{2}$WorldMind}\\[1pt]
  {\footnotesize $^{3}$Universiti Tunku Abdul Rahman}
}

\usepackage{wrapfig}

\begin{document}
\maketitle

\begin{abstract}
EEG-to-motion decoding remains challenging because noisy scalp EEG is difficult to map directly to continuous kinematics. We present NeuroMotion, a framework that aligns EEG representations to discrete motion primitives for continuous motion decoding across heterogeneous paradigms, including full-body free movement, lower-limb treadmill locomotion, and upper-limb freewill reaching and grasping. A VQ-based motion prior discretizes continuous motion into reusable primitives that serve as both alignment prototypes and intermediate prediction targets. Prototypical and supervised contrastive objectives align EEG embeddings to motion-derived prototypes and organize them according to primitive assignments. An E2M Transformer predicts motion-code sequences from the learned EEG representations, while a final geometric refinement stage improves continuous trajectory reconstruction through motion-level training. Under a last-session test protocol, NeuroMotion achieves the best mean reconstruction performance among the compared EEG decoding baselines across all three datasets, with the clearest gains on full-body motion in both reconstruction accuracy and distribution-level quality. Multi-subject ablations further support the contributions of EEG-to-primitive alignment, token prediction, and refinement. Code is available at \url{https://github.com/Liuyi2351433/NeuroAction}.
\end{abstract}

\section{Introduction}

Recent years have seen significant advances in brain-computer interfaces (BCIs), enabling applications ranging from communication restoration and neurorehabilitation to closed-loop neuromodulation and human-computer interaction. Among these applications, a central goal is to translate neural activity into actionable motor commands for external effectors, such as robotic arms, prosthetic limbs, exoskeletons, and humanoid agents. Achieving this goal requires robust EEG-to-motion decoding: inferring continuous human movement from non-invasive brain signals.

Despite steady progress, EEG-to-motion decoding remains highly challenging. Most existing methods are developed for specific motor paradigms, such as hand movement, reaching, or simplified gait cycles, and typically regress low-level kinematic variables directly from EEG~\citep{bradberry2010reconstructing}. While effective in constrained laboratory settings, such direct regression becomes difficult to scale across heterogeneous motor behaviors, long temporal horizons, and cross-session variations. A key reason is that EEG-to-motion decoding is not a strict one-to-one mapping. Scalp EEG provides noisy and spatially coarse measurements of population-level neural dynamics, rather than precise muscle-level commands. Neurophysiological studies further suggest that motor control is organized hierarchically, relying on internal forward models and modular motor primitives instead of explicitly specifying all joint trajectories~\citep{wolpert1998internal,mussaIvaldi2000primitives}. Optimal feedback control theory also suggests that neural control prioritizes task-relevant variables rather than exact reproduction of detailed kinematics~\citep{todorov2002optimal}. These observations motivate using motion-derived primitives as structured targets for EEG representation learning and as an intermediate representation for continuous motion reconstruction.

\begin{figure}[!t]
  \centering
  \includegraphics[width=1.0\linewidth]{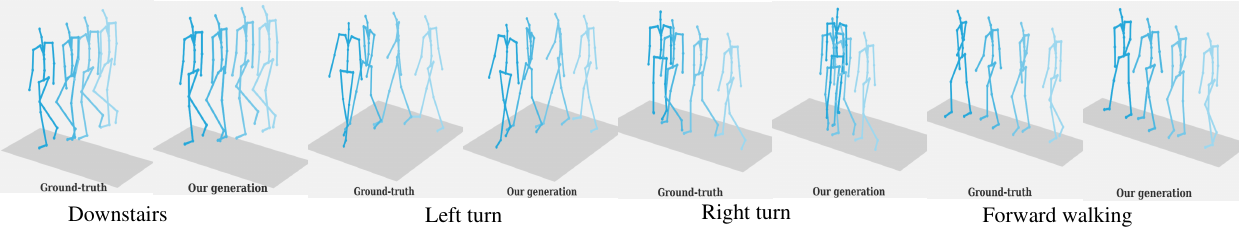}
  \vspace{-3mm}
\caption{Qualitative EEG-to-motion decoding examples on FBM (subject S10, session 10). We show four representative locomotion patterns: forward walking, downstairs, right turn, and left turn. Each example compares the ground-truth motion with the NeuroMotion-generated motion side by side. Each segment contains 40 frames, and we visualize one frame every 10 frames for clarity.}
  \vspace{-4mm}
  \label{fig:teaser}
\end{figure}

Meanwhile, EEG-motion study is moving from controlled single-task settings toward naturalistic and heterogeneous behaviors, including treadmill walking with closed-loop BCI control~\citep{he2018treadmill}, freewill reaching and grasping~\citep{thapa2025freewill}, and full-body locomotion across complex terrains~\citep{zhu2018fbm}. The shift requires that EEG-to-motion frameworks go beyond task-specific regressors and can be applied across upper-limb, lower-limb, and full-body motor paradigms.

Recent progress in motion generation and vision-language-action modeling suggests that continuous motion can be discretized into reusable tokens. Vector-quantized autoencoders have been widely used to tokenize high-dimensional human motion for Transformer-based generation~\citep{oord2017vqvae,guo2022tm2t,t2mgpt2023}, and VQ-VLA further shows that quantization is also useful for low-dimensional action trajectories. However, EEG-to-motion decoding requires a different use of motion quantization: EEG is noisy, subject-dependent, and not an explicit symbolic command, but it is temporally aligned with motion. We therefore use motion codes both as structured motion-primitive prototypes for EEG representation and as discrete intermediate targets for reconstruction.

To this end, we propose NeuroMotion, a framework that aligns EEG representations to discrete motion primitives for continuous motion decoding. NeuroMotion learns dataset-specific primitives with a VQ tokenizer, aligns EEG embeddings to their codebook representations through prototypical and supervised contrastive learning, predicts motion-code sequences with an E2M Transformer, and refines decoded trajectories through continuous motion-level supervision. We evaluate NeuroMotion on three EEG-motion datasets covering full-body movement, treadmill locomotion, and freewill reaching and grasping. Under a strict last-session test protocol, NeuroMotion achieves the best mean reconstruction performance among the compared EEG decoding baselines across all three datasets, with the clearest gains on full-body motion. Multi-subject ablations further support the contributions of EEG-to-primitive alignment, token prediction, and geometric refinement.

Our key contributions are summarized as follows:
\begin{itemize}[leftmargin=1em]
    \item We adopt EEG-to-primitive alignment, using motion-derived codebook prototypes and primitive assignments to structure continuous EEG representations via prototypical and contrastive learning.

    \item We develop NeuroMotion, a hierarchical EEG-to-motion framework that integrates motion tokenization, representation alignment, and motion-level refinement for continuous motion reconstruction.

    \item Across three heterogeneous EEG-motion datasets, NeuroMotion achieves the best mean reconstruction performance over baselines under a strict last-session test protocol, with the clearest gains on full-body motion and multi-subject ablations supporting alignment and staged decoding.
\end{itemize}


\begin{figure*}[t]
    \centering
    \includegraphics[width=\textwidth]{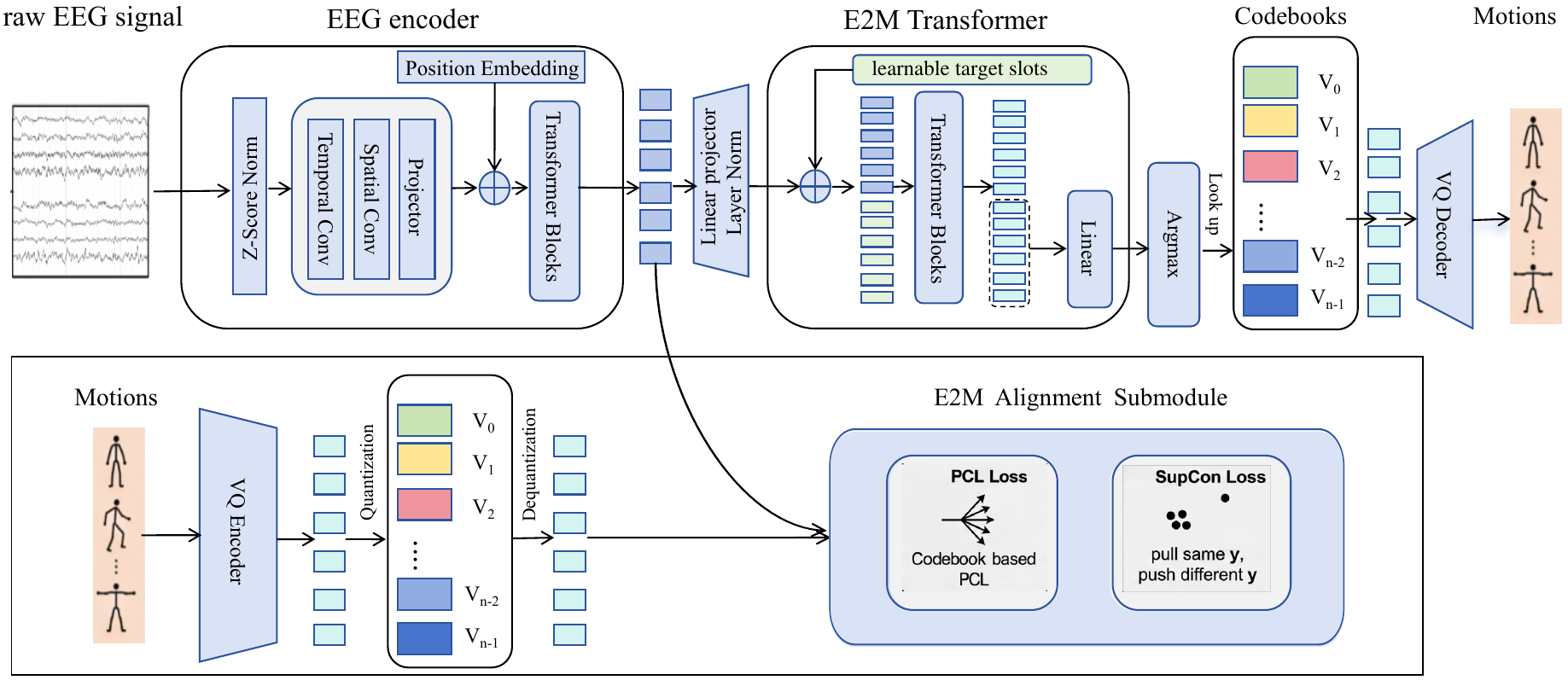}
    \vspace{-4mm}
    \caption{NeuroMotion architecture. EEG segments are encoded into continuous representations trained to align with discrete motion primitives learned by a VQ tokenizer. The E2M Transformer predicts motion-code sequences from these representations, and the motion decoder reconstructs continuous trajectories from the selected code vectors.}
    \label{fig:model_full}
\end{figure*}


\section{Method}
\subsection{Problem Formulation}
\label{sec:methods_formulation}
Let $X \in \mathbb{R}^{C \times T_e}$ be an EEG segment with $C$ electrodes and $T_e$ time samples, and let $Y \in \mathbb{R}^{T_m \times D}$ be its paired motion sequence, where $D$ depends on the target motion representation. NeuroMotion employs a discrete latent motion sequence $S=(s_1,\ldots,s_N)$, where each $s_i \in \{1,\ldots,K\}$ indexes one code from a learned motion codebook of size $K$. Each code corresponds to a short kinematic primitive in the motion manifold. We denote EEG-to-motion decoding as a latent generation process:
\begin{equation}
p(Y \mid X)
\approx
\sum_{S}
p_{\theta}(Y \mid S)\,p_{\phi}(S \mid X),
\label{eq:latent_factorization}
\end{equation}
where $p_{\phi}(S \mid X)$ predicts motion primitives from EEG, and $p_{\theta}(Y \mid S)$ reconstructs continuous motion from the selected primitives.

Under this formulation, EEG provides coarse neural evidence for selecting motion primitives, while the motion decoder synthesizes the final continuous trajectory from the learned motion prior. This yields a unified interface for heterogeneous EEG-motion settings with different output dimensions and temporal resolutions.

\subsection{Discrete Motion Prior}
\label{sec:methods_vqvae}

We first train a lightweight CNN-based VQ-VAE as the motion tokenizer. Given a motion sequence $Y$, the encoder maps it to latent vectors $Z=(z_1,\ldots,z_N)$, where $z_i\in\mathbb{R}^{d_c}$. A codebook $\mathcal{C}=\{c_1,\ldots,c_K\}$ defines a discrete set of motion primitives. Each latent vector is assigned to its nearest codebook entry, i.e., $s_i=\arg\min_k\|z_i-c_k\|_2^2$, and the quantized latent is $\hat{z}_i=c_{s_i}$.

The VQ decoder reconstructs the continuous motion sequence from the quantized latent sequence $\hat{Z}=(\hat{z}_1,\ldots,\hat{z}_N)$. The tokenizer is trained with reconstruction and commitment losses:
\begin{equation}
\hat{Y}=D_{\mathrm{VQ}}(\hat{Z}),
\qquad
\mathcal{L}_{\mathrm{VQ}}
=
\|Y-\hat{Y}\|_1
+
\beta
\|Z-\operatorname{sg}[\hat{Z}]\|_2^2,
\end{equation}
where $\operatorname{sg}[\cdot]$ is stop-gradient. The codebook is updated using EMA statistics of assigned encoder outputs, with dead-code reset to mitigate codebook collapse. After training, the VQ decoder and codebook define a discrete motion-prior space for EEG decoding.

\subsection{EEG Encoder}
\label{sec:methods_eeg_encoder}

Given a batch of EEG segments $X \in \mathbb{R}^{B \times C \times T}$, the EEG encoder produces token-level neural embeddings $H=f_{\mathrm{EEG}}(X)\in\mathbb{R}^{B \times N_e \times d_e}$. The EEG token count $N_e$ is matched to the motion-token count $N_m$; we use $N_e=N_m=N$ in the alignment and loss equations. Each EEG token corresponds to a motion-primitive position, and unmasked encoder attention gives it full-window context. The EEG encoder first applies channel-wise z-score normalization, then uses a lightweight convolutional patch embedding to form temporal EEG tokens, followed by a Transformer encoder~\citep{vaswani2017attention} to model temporal dependencies:
\begin{equation}
\tilde{X}=\operatorname{ZScore}(X),
\qquad
U=\operatorname{PatchEmbed}(\tilde{X}),
\qquad
H=\operatorname{TransEnc}(U).
\end{equation}

\subsection{EEG-to-Motion Token Decoder}
\label{sec:methods_e2m}

The EEG-to-motion decoder predicts motion codes from EEG embeddings $H$. An adapter maps $H$ to $H^{\mathrm{e2m}}\in\mathbb{R}^{B\times N_e\times d_m}$, where $d_m$ is the E2M hidden width. Learnable target slots $Q\in\mathbb{R}^{N_m\times d_m}$ are expanded across the batch to $Q^B\in\mathbb{R}^{B\times N_m\times d_m}$. Concatenation is along the sequence axis:
\begin{equation}
\begin{aligned}
U^{\mathrm{e2m}}&=[H^{\mathrm{e2m}};Q^B],\qquad
Z^{\mathrm{e2m}}=\mathcal{T}_{\mathrm{e2m}}(U^{\mathrm{e2m}}),\\
P_{b,i,:}&=\operatorname{Linear}(Z^{\mathrm{e2m}}_{b,N_e+i,:}),\qquad i=1,\ldots,N_m.
\end{aligned}
\label{eq:e2m_logits}
\end{equation}
Here $Z^{\mathrm{e2m}}\in\mathbb{R}^{B\times(N_e+N_m)\times d_m}$ and $P\in\mathbb{R}^{B\times N_m\times K}$, with the motion-token logits read from the target-slot positions.

The E2M Transformer applies a causal attention mask over the full ordered input sequence $[H^{\mathrm{e2m}}_1,\ldots,H^{\mathrm{e2m}}_{N_e},Q_1,\ldots,Q_{N_m}]$. Slot $Q_j$ can attend to all EEG tokens and slots $Q_1,\ldots,Q_j$, but not future slots. Since the upstream EEG encoder is bidirectional, each conditioning token already contains full-window context. All logits are produced in one forward pass: slots are dependent, but predicted IDs are never fed back as inputs. Appendix~\ref{app:e2m_transformer_arch} gives the mask and tensor shapes.

The predicted token is selected by maximum logit, and the token prediction loss is the cross-entropy against the ground-truth motion tokens:
\begin{equation}
\hat{s}_{b,i}=\arg\max_{k}P_{b,i,k},
\qquad
\mathcal{L}_{\mathrm{tok}}
=
-\frac{1}{BN}
\sum_{b=1}^{B}
\sum_{i=1}^{N}
\log
\frac{
\exp(P_{b,i,s_{b,i}})
}{
\sum_{k=1}^{K}
\exp(P_{b,i,k})
}.
\label{eq:token_loss}
\end{equation}
This loss trains the E2M decoder to generate the discrete motion-token sequence obtained from the paired ground-truth motion.

\subsection{EEG-to-Primitive Alignment}
\label{sec:methods_alignment}

In addition to token prediction, we align continuous EEG embeddings to motion-codebook prototypes and organize them according to the primitive assignments of paired motion. Let $h_{b,i}$ denote the EEG embedding at token position $i$ of sample $b$, and let $s_{b,i}$ be the corresponding ground-truth motion token obtained from the frozen motion tokenizer. The configured alignment width satisfies $d_e=d_c$, so the EEG embeddings and motion-code vectors can be compared directly. We normalize EEG embeddings and codebook vectors, and define the codebook-based prototypical contrastive loss as:
\begin{equation}
\bar{h}_{b,i}=\frac{h_{b,i}}{\|h_{b,i}\|_2},
\qquad
\bar{c}_{k}=\frac{c_k}{\|c_k\|_2},
\qquad
\mathcal{L}_{\mathrm{pcl}}
=
-\frac{1}{BN}
\sum_{b=1}^{B}
\sum_{i=1}^{N}
\log
\frac{
\exp(\bar{h}_{b,i}^{\top}\bar{c}_{s_{b,i}}/\tau_c)
}{
\sum_{k=1}^{K}
\exp(\bar{h}_{b,i}^{\top}\bar{c}_{k}/\tau_c)
}.
\label{eq:pcl_loss}
\end{equation}

We further apply supervised contrastive learning over EEG tokens using the same motion-token labels. Let $m=(b,i)$ index each EEG token instance, with embedding $\bar{h}_m$ and label $s_m$. For each anchor $m$, the positive set is $\mathcal{P}(m)=\{n\neq m \mid s_n=s_m\}$, and valid anchors are collected in $\mathcal{V}=\{m:|\mathcal{P}(m)|>0\}$. The supervised contrastive loss is:
\begin{equation}
\mathcal{L}_{\mathrm{supcon}}
=
-\frac{1}{|\mathcal{V}|}
\sum_{m\in\mathcal{V}}
\frac{1}{|\mathcal{P}(m)|}
\sum_{p\in\mathcal{P}(m)}
\log
\frac{
\exp(\bar{h}_m^{\top}\bar{h}_p/\tau_s)
}{
\sum_{n\neq m}
\exp(\bar{h}_m^{\top}\bar{h}_n/\tau_s)
}.
\label{eq:supcon_loss}
\end{equation}

The PCL objective aligns each EEG embedding to the motion-code prototype indexed by its paired motion token. SupCon further encourages EEG embeddings sharing the same primitive label to form compact and separable clusters. Together, these objectives structure the EEG representation space for subsequent motion-token prediction.

\subsection{Differentiable Motion-Level Refinement}
\label{sec:methods_motion_refinement}

After EEG-to-motion token generation, NeuroMotion performs a final geometric refinement stage to improve continuous trajectory accuracy. Token-level supervision encourages correct primitive prediction, but it does not directly optimize the decoded motion sequence. We therefore introduce a differentiable decoding path from token logits to continuous motion: given token logits $P \in \mathbb{R}^{B \times N \times K}$, we use straight-through Gumbel-Softmax to obtain differentiable hard token selections:
\begin{equation}
G_{b,i}
=
\operatorname{GumbelSoftmax}(P_{b,i}; \tau, \mathrm{hard=True}),
\qquad
G_{b,i}\in\mathbb{R}^{K}.
\end{equation}
The forward selections are one-hot, while the backward pass uses relaxed gradients. Thus $G\in\mathbb{R}^{B\times N_m\times K}$ retrieves code embeddings of shape $B\times N_m\times d_c$. The selected motion-code embedding is computed as $\tilde{z}_{b,i}=G_{b,i}\mathcal{C}$, where $\mathcal{C}\in\mathbb{R}^{K\times d_c}$ is the motion codebook. The VQ decoder then reconstructs the continuous motion $\tilde{Y}=D_{\mathrm{VQ}}(\tilde{Z})$. The refinement stage is optimized only with a frame-level MSE loss:
\begin{equation}
\mathcal{L}_{\mathrm{GR}}
=
\operatorname{MSE}(\tilde{Y},Y).
\end{equation}
 During this stage, gradients flow through the straight-through token-selection path and the VQ decoder, allowing the EEG encoder, E2M Transformer, and motion decoder to be refined jointly. This completes the coarse-to-fine hierarchy by converting discrete primitive predictions into more accurate continuous motion trajectories.

\subsection{Training Procedure}
\label{sec:methods_training}

NeuroMotion is trained in three stages. In Stage 1, we train the motion VQ-VAE on motion sequences only, producing a dataset-specific motion codebook and decoder that define the discrete motion-prior space. Each paired motion sequence is then quantized into a token sequence using nearest-neighbor codebook assignment.

In Stage 2, the motion tokenizer is frozen, and the EEG encoder and E2M Transformer are trained to predict motion tokens from EEG. The training objective combines token prediction with codebook-based PCL and supervised contrastive alignment:
\begin{equation}
\mathcal{L}_{\mathrm{stage2}}
=
\mathcal{L}_{\mathrm{tok}}
+
\lambda_{\mathrm{pcl}}\mathcal{L}_{\mathrm{pcl}}
+
\lambda_{\mathrm{supcon}}\mathcal{L}_{\mathrm{supcon}}.
\end{equation}

In Stage 3, we initialize from the Stage 2 model and apply the geometric refinement objective in Sec.~\ref{sec:methods_motion_refinement}. The VQ encoder and codebook remain fixed, while the EEG encoder, E2M Transformer, and VQ decoder are updated using continuous motion-level supervision. Overall, training proceeds from learning discrete motion primitives to EEG-to-primitive alignment and token prediction, followed by continuous motion-level refinement.

\subsection{Inference}
\label{sec:methods_inference}

At inference time, only EEG is required. Given an EEG segment, the EEG encoder first extracts neural embeddings, and the E2M Transformer predicts a discrete motion-token sequence. Hard argmax token IDs are mapped to their frozen codebook embeddings and decoded by the VQ decoder into the final continuous motion trajectory. Geometric refinement is a training stage, so inference requires no iterative motion optimization.

\section{Experiments}
\label{sec:experiments}

\subsection{Experimental Settings}
\label{sec:exp_setup}

\noindent\textbf{Datasets.}
We evaluate NeuroMotion on three public EEG-motion datasets spanning heterogeneous motor paradigms: full-body multi-terrain locomotion (FBM)~\citep{zhu2018fbm}, treadmill lower-limb locomotion (TM)~\citep{he2018treadmill}, and freewill reaching and grasping (FR)~\citep{thapa2025freewill}.
For FBM, we use subjects \texttt{s02}--\texttt{s10}, excluding \texttt{s01} due to synchronization issues, and 59 EEG channels after removing \texttt{FCz}; for TM, we evaluate six subjects using all 60 channels, excluding \texttt{S03} and \texttt{S04} due to data quality issues; and for FR, we use all 31 channels from \texttt{s05}, \texttt{s07}, \texttt{s13}, \texttt{s15}, and \texttt{s22}, the five subjects with complete three-session recordings.
Dataset acquisition schedules and the distinction between cross-day and session-disjoint evaluation are detailed in Appendix~\ref{app:dataset_protocols}.

\noindent\textbf{Motion representation.}
We use dataset-specific motion targets according to the available kinematic signals.
For FBM, we first select 22 joints from the original 23-joint representation by removing the L3 joint and reorder them to match the SMPL-style HumanML3D joint convention; the resulting motion is then converted to the 263-D HumanML3D-style representation.
For TM, the target is a 6-D lower-limb joint-angle sequence measured from bilateral hip, knee, and ankle goniometers.
For FR, the target is the synchronized three-axis wrist acceleration sequence.

\noindent\textbf{Data split.}
We use a strict session-wise split to avoid temporal leakage.
For each subject, the last available session is used for testing, while earlier sessions are used for training.
The motion VQ-VAE is trained separately for each dataset using only kinematic data from the training sessions; held-out EEG-to-motion test sessions are never used for VQ-VAE training.

\noindent\textbf{Training details.}
Across datasets, kinematic signals are low-pass filtered at 5\,Hz, downsampled to 20\,Hz, and temporally aligned with EEG, while EEG is band-pass filtered at 0.1--48\,Hz, re-referenced using common average reference (CAR)~\citep{delorme2004eeglab}, and downsampled to 100\,Hz.
Paired signals are segmented into 2\,s windows with a 0.5\,s stride.
TM and FR motion targets are min-max normalized to $[0,1]$, Channel-wise z-score normalization is applied to EEG as the first operation of the EEG encoder.
The default VQ codebook sizes are 32/16/12 and the motion-token counts are 10/10/5 for FBM/TM/FR, respectively.
Dataset-specific model widths, optimizer settings, learning-rate schedules, iteration budgets, and training costs are provided in Appendix~\ref{app:implementation}.

\providecommand{\rev}[1]{{#1}}
\pdfstringdefDisableCommands{\def\rev#1{#1}}
\newcommand{\mstd}[2]{$#1_{\pm #2}$}
\newcommand{\bmstd}[2]{$\mathbf{#1}_{\pm #2}$}
\newcommand{\umstd}[2]{$\underline{#1}_{\pm #2}$}

\begin{table}[t]
\centering
\caption{
Comparison with adapted EEG-to-motion baselines under the last-session test protocol.
Values are cohort mean $\pm$ between-subject standard deviation.
FBM PCC/$R^2$/RMSE evaluate 30-D end-effector positions; FID and Diversity use full 263-D motion.
GT denotes real-motion diversity.
\rev{Diversity is reported without ranking and interpreted relative to GT.}
\rev{Parameters (M) below method names follow FBM/FR/TM order.}
Subject-level results and paired tests are reported in Appendix~\ref{app:cohort_statistics}.
}
\label{tab:sota_combined}

\setlength{\tabcolsep}{2.6pt}
\renewcommand{\arraystretch}{0.95}
\small

\resizebox{\linewidth}{!}{%
\begin{tabular}{
@{}lccccc
@{\hspace{5pt}\vrule\hspace{5pt}}
ccc
@{\hspace{5pt}\vrule\hspace{5pt}}
ccc@{}
}
\toprule
& \multicolumn{5}{c}{\textbf{FBM} ($n=9$)}
& \multicolumn{3}{c}{\textbf{FR} ($n=5$)}
& \multicolumn{3}{c}{\textbf{TM} ($n=6$)} \\
\cmidrule(lr){2-6}
\cmidrule(lr){7-9}
\cmidrule(lr){10-12}

Method
& PCC$\uparrow$
& $R^2\uparrow$
& RMSE$\downarrow$
& FID$\downarrow$
& DIV$\rightarrow$
& PCC$\uparrow$
& $R^2\uparrow$
& RMSE$\downarrow$
& PCC$\uparrow$
& $R^2\uparrow$
& RMSE$\downarrow$ \\
\midrule

GT
& -- & -- & -- & --
& \mstd{5.86}{2.93}
& -- & -- & --
& -- & -- & -- \\

\shortstack[l]{\rev{LR}\\[-1pt]{\scriptsize \rev{(124 / 0.74 / 2.9)}}}
& \rev{\mstd{0.218}{0.158}}
& \rev{\mstd{-1.374}{1.066}}
& \rev{\mstd{1.215}{0.469}}
& \rev{\mstd{16.70}{19.42}}
& \rev{\mstd{5.89}{2.51}}
& \rev{\mstd{0.298}{0.164}}
& \rev{\mstd{-0.611}{0.747}}
& \rev{\mstd{0.343}{0.102}}
& \rev{\mstd{0.183}{0.125}}
& \rev{\mstd{-1.720}{2.290}}
& \rev{\mstd{0.719}{0.254}} \\

\shortstack[l]{\rev{RR}\\[-1pt]{\scriptsize \rev{(124 / 0.74 / 2.9)}}}
& \rev{\mstd{0.219}{0.158}}
& \rev{\mstd{-1.359}{1.052}}
& \rev{\mstd{1.211}{0.463}}
& \rev{\mstd{16.64}{19.39}}
& \rev{\mstd{5.89}{2.51}}
& \rev{\mstd{0.285}{0.154}}
& \rev{\mstd{-0.678}{0.776}}
& \rev{\mstd{0.314}{0.072}}
& \rev{\mstd{0.166}{0.130}}
& \rev{\mstd{-1.759}{2.403}}
& \rev{\mstd{0.678}{0.238}} \\

\shortstack[l]{\rev{UKF}\\[-1pt]{\scriptsize \rev{(0.26 / 0.02 / 0.06)}}}
& \rev{\mstd{0.182}{0.149}}
& \rev{\mstd{-0.613}{1.051}}
& \rev{\mstd{1.041}{0.570}}
& \rev{\mstd{14.76}{18.54}}
& \rev{\mstd{4.80}{2.71}}
& \rev{\mstd{0.193}{0.131}}
& \rev{\mstd{-2.614}{3.393}}
& \rev{\mstd{0.493}{0.222}}
& \rev{\mstd{0.183}{0.142}}
& \rev{\mstd{-2.213}{3.210}}
& \rev{\mstd{0.731}{0.354}} \\

\shortstack[l]{MTRT\\[-1pt]{\scriptsize \rev{(17 / 17 / 17)}}}
& \mstd{0.033}{0.012}
& \mstd{-0.223}{0.216}
& \mstd{0.898}{0.251}
& \mstd{33.3}{27.3}
& \mstd{3.43}{1.81}
& \mstd{0.505}{0.242}
& \mstd{0.111}{0.365}
& \mstd{0.254}{0.044}
& \mstd{0.071}{0.154}
& \mstd{-0.058}{0.107}
& \mstd{0.468}{0.025} \\

\shortstack[l]{DeepConvNet\\[-1pt]{\scriptsize \rev{(22 / 0.32 / 0.42)}}}
& \mstd{0.032}{0.041}
& \mstd{-0.262}{0.154}
& \mstd{0.951}{0.231}
& \mstd{62.5}{33.7}
& \mstd{2.92}{1.60}
& \mstd{0.506}{0.232}
& \mstd{-0.015}{0.468}
& \mstd{0.264}{0.040}
& \mstd{0.389}{0.232}
& \mstd{0.024}{0.239}
& \mstd{0.443}{0.055} \\

\shortstack[l]{STTransformer\\[-1pt]{\scriptsize \rev{(25 / 2.7 / 2.7)}}}
& \mstd{0.017}{0.008}
& \mstd{-0.365}{0.244}
& \mstd{0.941}{0.257}
& \mstd{38.1}{33.3}
& \mstd{2.52}{2.53}
& \mstd{0.489}{0.215}
& \mstd{-0.066}{0.533}
& \mstd{0.272}{0.051}
& \mstd{0.377}{0.229}
& \mstd{0.063}{0.123}
& \mstd{0.434}{0.047} \\

\shortstack[l]{NeuroDyGait\\[-1pt]{\scriptsize \rev{(5.7 / 2.4 / 2.4)}}}
& \mstd{0.049}{0.028}
& \mstd{-0.212}{0.203}
& \mstd{0.897}{0.251}
& \mstd{65.2}{38.2}
& \mstd{3.94}{1.34}
& \mstd{0.607}{0.164}
& \umstd{0.162}{0.370}
& \umstd{0.240}{0.033}
& \mstd{0.450}{0.165}
& \umstd{0.127}{0.134}
& \umstd{0.418}{0.039} \\

\shortstack[l]{EEGConformer\\[-1pt]{\scriptsize \rev{(25 / 0.86 / 0.93)}}}
& \mstd{0.026}{0.013}
& \mstd{-0.307}{0.133}
& \mstd{0.930}{0.233}
& \mstd{22.2}{23.4}
& \mstd{2.14}{2.12}
& \mstd{0.474}{0.235}
& \mstd{-0.154}{0.574}
& \mstd{0.280}{0.048}
& \mstd{0.333}{0.241}
& \mstd{0.032}{0.129}
& \mstd{0.443}{0.041} \\

\shortstack[l]{ContraWR\\[-1pt]{\scriptsize \rev{(24 / 1.6 / 1.7)}}}
& \mstd{0.026}{0.031}
& \mstd{-0.261}{0.145}
& \mstd{0.904}{0.229}
& \mstd{56.9}{28.4}
& \mstd{2.82}{1.56}
& \mstd{0.513}{0.306}
& \mstd{-0.006}{0.481}
& \mstd{0.266}{0.050}
& \mstd{-0.006}{0.015}
& \mstd{-0.058}{0.089}
& \mstd{0.470}{0.028} \\

\shortstack[l]{Ours-stage2\\[-1pt]{\scriptsize \rev{(89 / 24 / 22)}}}
& \umstd{0.412}{0.241}
& \umstd{-0.117}{0.392}
& \umstd{0.821}{0.245}
& \bmstd{3.17}{2.04}
& \rev{\mstd{4.48}{2.90}}
& \umstd{0.643}{0.146}
& \mstd{0.094}{0.525}
& \mstd{0.247}{0.052}
& \umstd{0.459}{0.224}
& \mstd{-0.102}{0.308}
& \mstd{0.463}{0.077} \\

\shortstack[l]{Ours-stage3\\[-1pt]{\scriptsize \rev{(89 / 24 / 22)}}}
& \bmstd{0.522}{0.238}
& \bmstd{0.110}{0.354}
& \bmstd{0.718}{0.229}
& \umstd{4.14}{4.04}
& \rev{\mstd{4.36}{2.89}}
& \bmstd{0.657}{0.152}
& \bmstd{0.250}{0.340}
& \bmstd{0.230}{0.038}
& \bmstd{0.485}{0.211}
& \bmstd{0.184}{0.148}
& \bmstd{0.404}{0.052} \\

\bottomrule
\end{tabular}%
}
\end{table}

\subsection{\rev{Comparison with Adapted EEG-to-Motion Baselines}}
\label{sec:exp_sota}

\noindent\textbf{Baselines.}
Since EEG-to-motion decoding has been studied under different motor paradigms and output formats, we compare NeuroMotion with representative baselines including the upper-limb decoder MTRT~\citep{zhang2024mtrt}, the gait-specific NeuroDyGait (DY)~\citep{fu2025neurodygait}, and general-purpose EEG models EEGConformer (EC)~\citep{song2023eegconformer}, STTransformer (ST)\rev{~\citep{song2021sttransformer}}, DeepConvNet (DCN)~\citep{schirrmeister2017deep}, and ContraWR~\citep{banville2023contrawr}.
\rev{We additionally compare linear regression (LR), ridge regression (RR), and an unscented Kalman filter (UKF) on all three datasets~\citep{nakagome2020comparison,luu2016unscented}.}
All methods use the same dataset splits, preprocessing, and prediction targets; detailed baseline adaptation and training controls are provided in Appendix~\ref{app:baseline_adaptation}.
Table~\ref{tab:sota_combined} reports cohort-level mean$\pm$std performance.

\begin{wrapfigure}{r}{0.55\linewidth}
    \centering
    \setlength{\abovecaptionskip}{1pt} 
    \setlength{\belowcaptionskip}{-10pt} 

    \includegraphics[width=\linewidth]{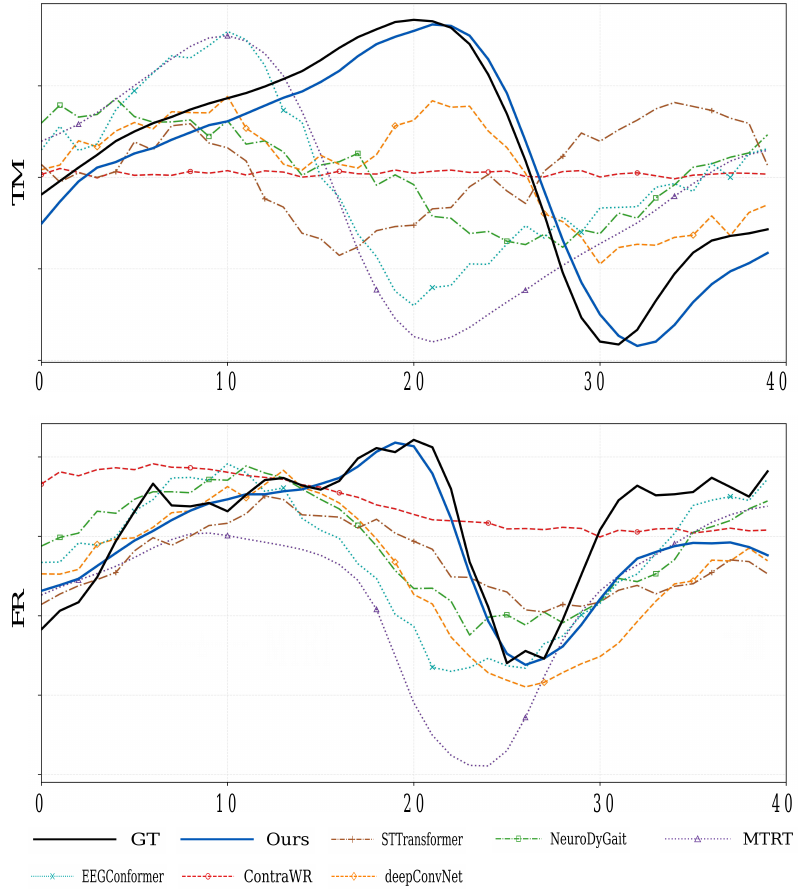}
    \caption{
    Comparison on TM (\textcolor{black}{left hip} $z$-axis
    \textcolor{black}{angle}) and FR
    (\textcolor{black}{$X$-axis wrist acceleration}).
    \textcolor{black}{The top and bottom panels show representative
    40-frame test windows from TM \texttt{S07} and FR \texttt{S05},
    respectively, both from held-out session~03.
    Black and blue curves denote ground truth and our predictions,
    respectively; the remaining curves show baseline predictions.}
    }
    \label{fig:tm_fr_vis}
\end{wrapfigure}

\begin{figure}[!t]
    \centering
    \includegraphics[width=\linewidth]{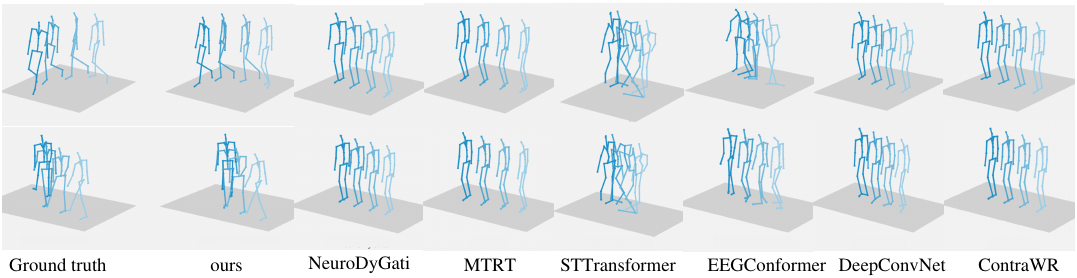}
    \vspace{-7mm}
    \caption{Qualitative comparison on FBM. We visualize decoded full-body motion from NeuroMotion and representative baselines against the ground truth. Compared with baselines that often produce over-smoothed or disordered poses, NeuroMotion better preserves the locomotion pattern, articulated body structure, and overall motion coherence.}
    \vspace{-2mm}
    \label{fig:fbm_vis}
\end{figure}

\noindent\textbf{Overall performance and uncertainty.}
Figures~\ref{fig:fbm_vis} and~\ref{fig:tm_fr_vis} complement the aggregate metrics with representative qualitative reconstructions across the three datasets. NeuroMotion captures the dominant temporal structure of full-body motion on FBM and follows the major joint-angle and wrist-acceleration trends on TM and FR, while fine-scale deviations remain visible.

Table~\ref{tab:sota_combined} shows that NeuroMotion Stage 3 attains the highest mean reconstruction performance among the compared methods on all three datasets.
\rev{On FBM, PCC improves from RR's $0.219\pm0.158$ to $0.522\pm0.238$; $R^2$/RMSE improve from NeuroDyGait's $-0.212\pm0.203$/$0.897\pm0.251$ to $0.110\pm0.354$/$0.718\pm0.229$, respectively.}
Positive mean $R^2$ implies non-trivial variance capture, although \rev{absolute reconstruction accuracy remains limited and between-subject variability is substantial}.

The gains on TM and FR are smaller and accompanied by substantial between-subject variability, especially given the limited cohort sizes.
Complete subject-level results, improved-subject counts, and paired statistical comparisons are provided in Appendix~\ref{app:cohort_statistics}.
Overall, the performance gains are clearest on FBM, while conclusions on TM and FR remain more limited.

\rev{The added classical decoders attain mean PCC/$R^2$/RMSE of $0.218/-1.374/1.215$ (LR), $0.219/-1.359/1.211$ (RR), and $0.182/-0.613/1.041$ (UKF) on FBM. NeuroMotion retains the highest mean PCC and $R^2$ and the lowest mean RMSE on all three datasets. Complete subject-wise results for the added decoders are provided in Appendix~\ref{app:subjectwise_baselines}.}

\setlength{\intextsep}{6pt} 

\noindent\textbf{Hard-token decoding and refinement.}
Stage 2 already provides effective motion reconstruction from predicted hard token IDs alone, achieving PCC/$R^2$/RMSE of $0.412/-0.117/0.821$ on FBM, $0.459/-0.102/0.463$ on TM, and $0.643/0.094/0.247$ on FR. This shows that the discrete motion-token stage captures useful motion structure before any continuous refinement. Stage 3 then further improves the reconstructed trajectories: all three metrics improve for every FBM and TM subject, while on FR PCC improves for 5/5 subjects and $R^2$/RMSE for 4/5, with S07 unchanged on the latter two metrics. Together, these results support complementary roles for the two stages, with Stage 2 providing coarse motion structure and Stage 3 refining it in continuous kinematic space.

\noindent\textbf{Distributional quality.}
\rev{On FBM, Stage 2 has the lowest FID ($3.169\pm2.037$), with Diversityt of $4.480$ relative to the real-motion reference ($5.86$). Diversity is reported as a reference-based distributional descriptor without ranking.}
Stage 3 improves paired reconstruction while FID increases to $4.139\pm4.036$.
The default Stage-3 objective supervises 30-D end effectors, whereas FID measures full-263-D sequences using the pretrained motion evaluator of Guo et al.~\citet{guo2022diversehumanmotion}, which is widely used for text-to-motion evaluation.
A controlled full-263-D refinement objective does not resolve this trade-off (Appendix~\ref{app:refinement_scope}).
Given evaluator-domain mismatch and substantial subject variability, FID is complementary evidence rather than a direct measure of EEG-conditioned trajectory accuracy.

\subsection{Ablation Study}
\label{sec:exp_ablation}

We conduct multi-subject component ablations on four subjects per dataset: FBM S02/S07/S09/S10, TM S01/S02/S06/S07, and FR S05/S07/S13/S22. Table~\ref{tab:ablation_three_datasets} reports mean $\pm$ between-subject standard deviation. Removing either alignment objective or both degrades mean performance across all three datasets, supporting the contribution of codebook-guided EEG alignment. Skipping Stage 2 causes the largest overall degradation, supporting the importance of learning the discrete intermediate task before motion-level optimization.

\setlength{\intextsep}{6pt} 

\begin{wrapfigure}{r}{0.5\textwidth}
    \centering
    \setlength{\abovecaptionskip}{1pt} 
    \setlength{\belowcaptionskip}{-10pt} 

    \includegraphics[width=\linewidth]{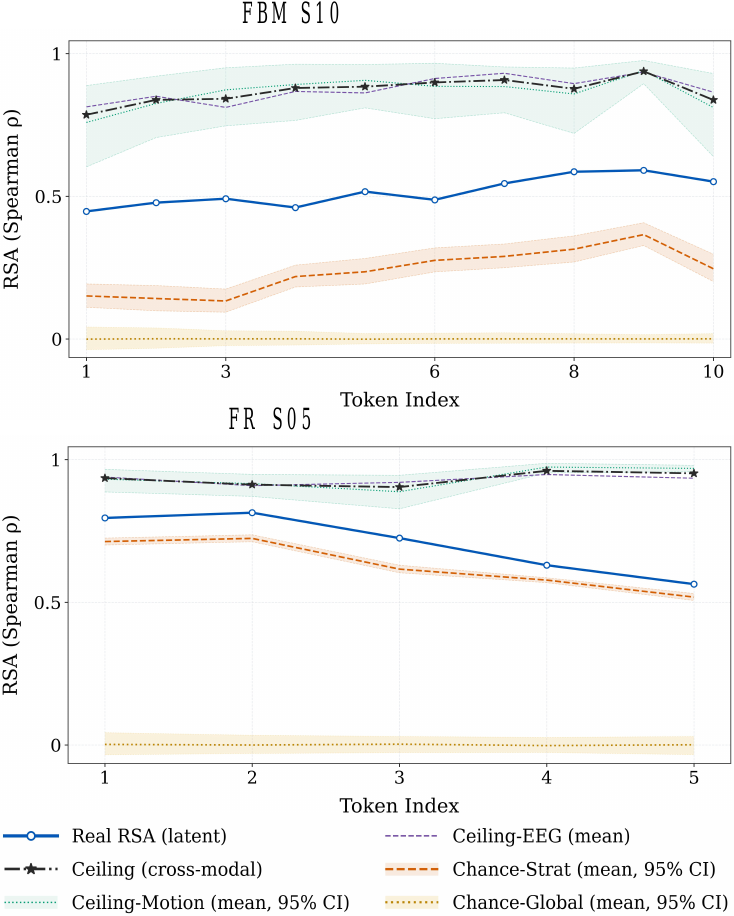}
    \caption{
    Token-level RSA on FBM \texttt{S10}
    \textcolor{black}{(session~11)} and FR
    \textcolor{black}{\texttt{S05} (session~03)}.
    \textcolor{black}{
    Blue curves show EEG--motion RSA using pre-quantized motion latents.
    The other curves show ceiling estimates and permutation-based
    chance controls.
    Shaded bands indicate 95\% confidence intervals for the motion
    ceiling and both chance baselines.
    }
    }
    \label{fig:rsa_token}
\end{wrapfigure}

\textit{w/o SupCon} removes supervised contrastive learning, \textit{w/o PCL} removes prototypical contrastive learning, and \textit{w/o Alignment} removes both. \textit{w/o E2M} removes the token-generation branch and cross-entropy supervision: EEG embeddings are trained using alignment losses alone, tokens are selected by nearest-codebook cosine similarity, and Stage 3 is omitted. Thus, this setting ablates the learned token-generation pathway rather than only the Transformer architecture. \textit{w/o Stage 2} retains the Stage-1 motion prior and E2M architecture, but skips token-prediction and alignment training and directly optimizes the EEG encoder, E2M module, and VQ decoder with motion-space MSE. \rev{\textit{Primitive classifier} removes both alignment objectives and E2M, predicts the motion-code label with a classifier on each EEG token, and reconstructs the trajectory using the same motion decoder. It uses the same four-subject subsets and performs worse than the full model and \textit{w/o Alignment} on all three cohort-mean reconstruction metrics across datasets (Table~\ref{tab:ablation_three_datasets}), supporting the contribution of alignment-guided temporal decoding beyond discretization alone.} The effect of Stage 3 is evaluated separately on the complete cohorts in Table~\ref{tab:sota_combined} and Appendix~\ref{app:cohort_statistics}.

\providecommand{\mstd}[2]{$#1^{\pm #2}$}
\providecommand{\bmstd}[2]{$\mathbf{#1}^{\boldsymbol{\pm #2}}$}
\providecommand{\umstd}[2]{$\underline{#1^{\pm #2}}$}

\begin{table}[tbp]
\centering
\caption{Component ablations on four subjects per dataset: FBM (S02/S07/S09/S10), TM (S01/S02/S06/S07), and FR (S05/S07/S13/S22). Entries are mean $\pm$ between-subject standard deviation. Ablation definitions are given in Sec.~\ref{sec:exp_ablation}.}
\label{tab:ablation_three_datasets}

\small
\setlength{\tabcolsep}{3.5pt}
\renewcommand{\arraystretch}{1.0}

\makebox[\linewidth][c]{%
\resizebox{1.0\linewidth}{!}{%
\begin{tabular}{
@{}l
ccc
@{\hspace{6pt}\vrule\hspace{6pt}}
ccc
@{\hspace{6pt}\vrule\hspace{6pt}}
ccc@{}
}
\toprule

& \multicolumn{3}{c}{\textbf{FBM}}
& \multicolumn{3}{c}{\textbf{FR}}
& \multicolumn{3}{c}{\textbf{TM}} \\
\cmidrule(lr){2-4}
\cmidrule(lr){5-7}
\cmidrule(lr){8-10}

Variant
& PCC$\uparrow$ & $R^2\uparrow$ & RMSE$\downarrow$
& PCC$\uparrow$ & $R^2\uparrow$ & RMSE$\downarrow$
& PCC$\uparrow$ & $R^2\uparrow$ & RMSE$\downarrow$ \\
\midrule

Full model
& \bmstd{0.584}{0.231}
& \bmstd{0.259}{0.377}
& \bmstd{0.746}{0.291}
& \bmstd{0.717}{0.087}
& \bmstd{0.350}{0.297}
& \bmstd{0.216}{0.024}
& \bmstd{0.522}{0.189}
& \bmstd{0.187}{0.098}
& \bmstd{0.404}{0.066} \\

w/o SupCon
& \umstd{0.373}{0.349}
& \umstd{0.163}{0.289}
& \umstd{0.836}{0.296}
& \mstd{0.612}{0.144}
& \mstd{0.174}{0.400}
& \mstd{0.242}{0.028}
& \mstd{0.338}{0.320}
& \umstd{0.062}{0.091}
& \umstd{0.434}{0.066} \\

w/o PCL
& \mstd{0.186}{0.329}
& \mstd{-0.015}{0.287}
& \mstd{0.940}{0.297}
& \umstd{0.648}{0.165}
& \umstd{0.283}{0.286}
& \umstd{0.229}{0.023}
& \mstd{0.188}{0.377}
& \mstd{0.012}{0.081}
& \mstd{0.446}{0.065} \\

w/o Alignment
& \mstd{0.259}{0.301}
& \mstd{0.033}{0.219}
& \mstd{0.920}{0.263}
& \mstd{0.596}{0.152}
& \mstd{0.150}{0.479}
& \mstd{0.244}{0.029}
& \mstd{0.164}{0.328}
& \mstd{-0.037}{0.028}
& \mstd{0.458}{0.041} \\

w/o E2M
& \mstd{0.317}{0.244}
& \mstd{-0.295}{0.476}
& \mstd{1.000}{0.341}
& \mstd{0.598}{0.151}
& \mstd{0.091}{0.447}
& \mstd{0.255}{0.037}
& \umstd{0.490}{0.194}
& \mstd{-0.075}{0.201}
& \mstd{0.461}{0.094} \\

\rev{Primitive classifier}
& \rev{\mstd{0.206}{0.240}}
& \rev{\mstd{-0.019}{0.237}}
& \rev{\mstd{0.972}{0.278}}
& \rev{\mstd{0.545}{0.148}}
& \rev{\mstd{0.061}{0.479}}
& \rev{\mstd{0.267}{0.018}}
& \rev{\mstd{0.151}{0.303}}
& \rev{\mstd{-0.097}{0.022}}
& \rev{\mstd{0.483}{0.045}} \\

w/o Stage 2
& \mstd{0.018}{0.022}
& \mstd{-0.396}{0.274}
& \mstd{1.078}{0.236}
& \mstd{0.170}{0.221}
& \mstd{-0.196}{0.155}
& \mstd{0.307}{0.053}
& \mstd{0.184}{0.137}
& \mstd{-0.165}{0.050}
& \mstd{0.492}{0.025} \\

\bottomrule
\end{tabular}%
}%
}
\end{table}

Appendix~\ref{app:extended_ablation} evaluates codebook capacity, temporal token granularity, alignment weighting, codebook fine-tuning, and refinement supervision scope. The experiments use explicitly specified subject subsets or the complete cohorts, with comparisons made only within matched experimental settings.

\begin{table}[tbp]
\centering
\caption{EEG input controls, averaged over FBM (S03/\rev{S04/S05/}S10), TM (\rev{S02/S05/}S07/S08), and FR (S05/\rev{S07/}S13\rev{/S15}). Reversal changes temporal orientation; shuffling destroys temporal order within each window. Stage 1 is reused and Stages 2--3 are retrained with the same settings.}
\label{tab:eeg_control}

\small
\setlength{\tabcolsep}{3.5pt}
\renewcommand{\arraystretch}{1.0}

\makebox[\linewidth][c]{%
\resizebox{0.96\linewidth}{!}{%
\begin{tabular}{
@{}lccc
@{\hspace{6pt}\vrule\hspace{6pt}}
ccc
@{\hspace{6pt}\vrule\hspace{6pt}}
ccc@{}
}
\toprule

& \multicolumn{3}{c}{\textbf{FBM}}
& \multicolumn{3}{c}{\textbf{FR}}
& \multicolumn{3}{c}{\textbf{TM}} \\

\cmidrule(lr){2-4}
\cmidrule(lr){5-7}
\cmidrule(lr){8-10}

Input
& PCC$\uparrow$
& $R^2\uparrow$
& RMSE$\downarrow$
& PCC$\uparrow$
& $R^2\uparrow$
& RMSE$\downarrow$
& PCC$\uparrow$
& $R^2\uparrow$
& RMSE$\downarrow$ \\

\midrule

Standard EEG
& \textbf{\rev{0.510}}
& \textbf{\rev{0.197}}
& \textbf{\rev{0.629}}
& \textbf{\rev{0.674}}
& \textbf{\rev{0.235}}
& \textbf{\rev{0.234}}
& \textbf{\rev{0.517}}
& \textbf{\rev{0.212}}
& \textbf{\rev{0.388}} \\

Reversed EEG
& \rev{0.408} & \rev{0.043} & \rev{0.742}
& \rev{0.645} & \rev{0.148} & \rev{0.274}
& \rev{0.395} & \rev{0.054} & \rev{0.449} \\

Shuffled EEG
& \rev{0.083} & \rev{-0.461} & \rev{1.046}
& \rev{0.511} & \rev{-0.162} & \rev{0.331}
& \rev{0.004} & \rev{-0.626} & \rev{0.627} \\

Zero EEG
& \rev{0.010} & \rev{-0.730} & \rev{1.150}
& 0.000 & \rev{-0.412} & \rev{0.385}
& \rev{0.003} & \rev{-0.699} & \rev{0.634} \\

\bottomrule
\end{tabular}%
}%
}

\end{table}

\noindent\textbf{EEG input controls.}
Table~\ref{tab:eeg_control} compares standard EEG with three input controls designed to probe different information sources in the decoding pipeline. Zero EEG evaluates how much performance can be attributed to the pretrained VQ-VAE motion prior alone, without informative neural input. Random temporal shuffling removes temporal order while preserving instantaneous multichannel values, testing how much decoding can rely on non-temporal EEG information. Temporal reversal provides an intermediate control that reverses temporal orientation while preserving local adjacency, Fourier magnitude, and cross-channel relations. For all controls, we reuse the Stage-1 motion prior and retrain Stages 2--3 under the same protocol.

For the order-destruction control, each 2-s window receives an independent random temporal permutation shared across channels. This preserves instantaneous multichannel vectors and marginal values while destroying temporal order and local continuity, with the EEG--motion window pairing unchanged. Shuffling reduces PCC to \rev{0.083} on FBM and \rev{0.004} on TM. On FR, PCC remains \rev{0.511}, but $R^2$ drops from \rev{0.235} to \rev{-0.162} and RMSE increases from \rev{0.234} to \rev{0.331}. Zero EEG performs poorly on all datasets. These controls indicate that ordered temporal structure contributes substantially to decoding on FBM and TM, while FR retains additional information in instantaneous multichannel structure despite a marked loss in reconstruction quality.

\setlength{\intextsep}{6pt} 

\begin{wrapfigure}{r}{0.50\textwidth}
    \centering
    \setlength{\abovecaptionskip}{4pt} 
    \setlength{\belowcaptionskip}{-10pt} 

    \includegraphics[width=\linewidth]{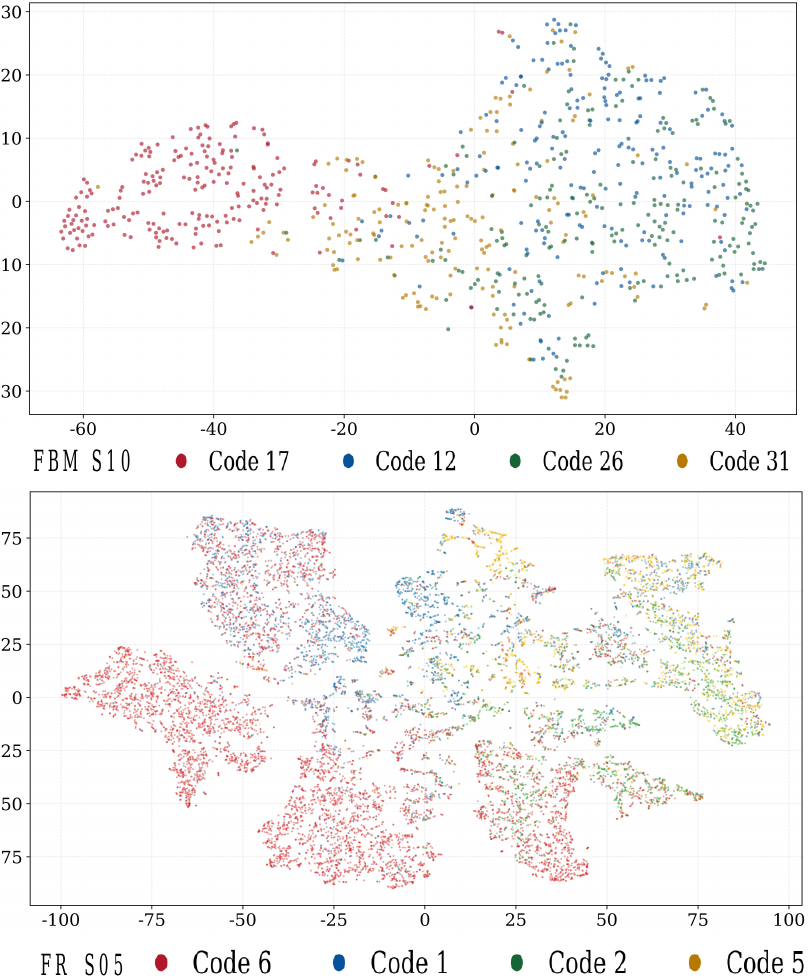}
    \caption{
    Token-level t-SNE visualization of EEG embeddings on FBM
    \texttt{S10} \textcolor{black}{(session~11)} and FR
    \texttt{S05} \textcolor{black}{(session~03)}.
    \textcolor{black}{
    Each point represents an EEG token, colored by the code ID assigned
    to its paired ground-truth motion token.
    The four most frequent motion codes in each test session are shown.
    Code IDs refer to the respective dataset-specific motion codebook.
    }
    }
    \label{fig:tsne_token}
\end{wrapfigure}

We further analyze whether the learned EEG representations are structurally aligned with motion primitives. We perform token-level RSA~\citep{kriegeskorte2008rsa} between EEG embeddings and motion representations by comparing their representational dissimilarity matrices (RDMs) at each token position. We report representative results on FBM \texttt{S10} and FR \texttt{S05}, with more details in Appendix~\ref{app:rsa_alignment}.

As shown in Fig.~\ref{fig:rsa_token}, the real EEG--motion RSA curves
remain above permutation-based chance baselines, supporting structural
correspondence between the learned EEG representations and the
motion-primitive space. The \textcolor{black}{latent RSA curve compares}
EEG embeddings with pre-quantized motion latents; chance curves are
obtained through \textcolor{black}{stratified and global permutation
controls}, and the \textcolor{black}{cross-modal} ceiling estimates the
attainable similarity given representational variability.

We also visualize token-level EEG embeddings with t-SNE on FBM
\texttt{S10} and FR \texttt{S05}.
\rev{Each panel contains EEG tokens from a single held-out test session
of the indicated subject.}
\textcolor{black}{The test sessions are 11 for FBM and 03 for FR.}
We select embeddings associated with the four most frequently used
motion codes \textcolor{black}{within each test session} and color each
point by its paired \textcolor{black}{ground-truth motion-code ID}.
As shown in Fig.~\ref{fig:tsne_token}, the embeddings exhibit
code-dependent organization, providing qualitative evidence of
correspondence with recurring motion primitives. Because the codebook
is learned solely from motion, these results reflect task-specific
EEG--motion correspondence rather than a direct physiological
categorization of motor states.

\section{Discussion}
\label{sec:discussion}
NeuroMotion uses discrete motion primitives as an intermediate representation between noisy EEG and continuous kinematics. Its main contribution is aligning continuous EEG embeddings to motion-derived prototypes and integrating this alignment with token prediction and motion-level refinement. A common formulation supports FBM, TM, and FR, using a dataset-specific motion tokenizer and subject-specific EEG-to-motion models.

Multi-subject ablations support the contributions of alignment and staged learning, while EEG input controls demonstrate sensitivity to ordered temporal dynamics. Frequency-band analysis (Appendix~\ref{app:filtering}) and representation analyses provide complementary diagnostics. Because the motion codebook is learned solely from kinematics, the resulting tokens should be interpreted as reusable motion primitives with task-specific EEG correspondence rather than independently established physiological motor states.

\textbf{Limitations.}
The study is limited by small subject cohorts and offline evaluation. \rev{The low mean $R^2$ on FBM and negative values for some subjects indicate limited absolute reconstruction accuracy and consistency across subjects.} Although TM and FR include cross-day testing, longer-term EEG nonstationarity remains unresolved, while FBM lacks verified day-level metadata. The discrete VQ motion prior may introduce quantization error and suppress fine-grained kinematic variation. Reported inference timings characterize model-side computation only and do not establish end-to-end closed-loop performance. Larger population studies, longer-term evaluation, and online validation remain important future directions.
\section{Conclusion}
\label{sec:conclusion}

We presented NeuroMotion, a framework that aligns EEG representations to discrete motion primitives and integrates motion-token prediction with geometric refinement for continuous motion decoding. Across three dataset-specific instantiations, the method improves mean reconstruction metrics over the compared baselines, with the strongest statistical evidence on FBM end-effector reconstruction. Multi-subject ablations, temporal controls, and representation analyses support structured EEG--motion alignment. Cross-day testing and offline profiling characterize current capabilities, while broader population studies and online closed-loop evaluation remain necessary.

\clearpage
\section*{AI Use Statement}

Generative AI tools were used solely for language polishing and improving the clarity and readability of the manuscript. They were not used to generate scientific ideas, experimental designs, code, data, analyses, results, or references. All technical content and scientific claims were produced and verified by the authors.

\section*{Ethics Statement}

This work uses only previously published public EEG--motion datasets and involves no new human-subject data collection. We follow the data-use conditions of the original datasets and use the recordings solely for offline research on EEG-to-motion decoding. Because EEG is a sensitive biosignal, practical deployment of such systems should require informed consent, appropriate privacy and data-security safeguards, and careful safety validation, particularly in closed-loop or assistive-control settings.

\section*{Reproducibility Statement}

We provide detailed experimental information in the appendix, including dataset acquisition and session protocols, preprocessing and normalization, motion representations, session-wise train/test splits, model architectures and tensor shapes, codebook and token configurations, optimization hyperparameters, baseline adaptation procedures, evaluation metrics, subject-level statistics, ablation settings, and inference-cost measurements. All reported experiments use explicitly specified subject subsets or complete cohorts, and held-out test sessions are excluded from motion-prior training. Code is available at \url{https://github.com/Liuyi2351433/NeuroAction}.

\clearpage

\bibliographystyle{iclr2027_conference}
\begingroup
\let\RoneOriginalBibitem\bibitem
\renewcommand{\bibitem}[2][]{%
  \color{black}%
  \ifcsname RoneNewReference@#2\endcsname\color{black}\fi
  \RoneOriginalBibitem[#1]{#2}}
\expandafter\def\csname RoneNewReference@nakagome2020comparison\endcsname{}
\expandafter\def\csname RoneNewReference@li2009ukf\endcsname{}
\bibliography{ref_nips}

@article{bradberry2010reconstructing,
  title        = {Reconstructing three-dimensional hand movements from noninvasive electroencephalographic signals},
  author       = {Bradberry, Trent J. and Gentili, Rodolphe J. and Contreras-Vidal, Jos{\'e} L.},
  journal      = {The Journal of Neuroscience},
  volume       = {30},
  number       = {9},
  pages        = {3432--3437},
  year         = {2010},
  doi          = {10.1523/JNEUROSCI.6107-09.2010}
}

@article{wolpert1998internal,
  title        = {Internal models in the cerebellum},
  author       = {Wolpert, Daniel M. and Miall, R. Chris and Kawato, Mitsuo},
  journal      = {Trends in Cognitive Sciences},
  volume       = {2},
  number       = {9},
  pages        = {338--347},
  year         = {1998},
  doi          = {10.1016/S1364-6613(98)01221-2}
}

@article{mussaIvaldi2000primitives,
  title        = {Modular features of motor control and learning},
  author       = {Mussa-Ivaldi, Ferdinando A.},
  journal      = {Current Opinion in Neurobiology},
  volume       = {9},
  number       = {6},
  pages        = {713--717},
  year         = {1999},
  doi          = {10.1016/S0959-4388(99)00029-X}
}

@article{todorov2002optimal,
  title        = {Optimal feedback control as a theory of motor coordination},
  author       = {Todorov, Emanuel and Jordan, Michael I.},
  journal      = {Nature Neuroscience},
  volume       = {5},
  number       = {11},
  pages        = {1226--1235},
  year         = {2002},
  doi          = {10.1038/nn963}
}

@article{zhu2018fbm,
  title        = {Full body mobile brain-body imaging data during unconstrained locomotion on stairs, ramps, and level ground},
  author       = {Brantley, Justin A. and Luu, Trieu Phat and Nakagome, Sho and Zhu, Fangshi and Contreras-Vidal, Jose L.},
  journal      = {Scientific Data},
  volume       = {5},
  pages        = {180133},
  year         = {2018},
  doi          = {10.1038/sdata.2018.133},
  url          = {https://www.nature.com/articles/sdata2018133.pdf}
}

@article{mohanta2025move4as,
  title        = {A Multimodal Dataset Addressing Motor Function in Autism},
  author       = {Paulo, Jo{\~a}o Ruivo and Sousa, Teresa and Perdiz, Jo{\~a}o and Pereira, Lara and Vasen, Mariette and Mouga, Susana and Pires, Gabriel and Castelo-Branco, Miguel},
  journal      = {Scientific Data},
  volume       = {12},
  pages        = {959},
  year         = {2025},
  doi          = {10.1038/s41597-025-05313-0},
  url          = {https://www.nature.com/articles/s41597-025-05313-0.pdf}
}

@inproceedings{oord2017vqvae,
  title        = {Neural Discrete Representation Learning},
  author       = {van den Oord, Aaron and Vinyals, Oriol and Kavukcuoglu, Koray},
  booktitle    = {Advances in Neural Information Processing Systems},
  volume       = {30},
  year         = {2017},
  url          = {https://arxiv.org/abs/1711.00937}
}

@inproceedings{guo2022tm2t,
  title        = {{TM2T}: Stochastic and Tokenized Modeling for the Reciprocal Generation of {3D} Human Motions and Texts},
  author       = {Guo, Chuan and Zuo, Xinxin and Wang, Sen and Cheng, Li},
  booktitle    = {Computer Vision -- ECCV 2022},
  series       = {Lecture Notes in Computer Science},
  volume       = {13695},
  pages        = {580--597},
  publisher    = {Springer},
  year         = {2022},
  doi          = {10.1007/978-3-031-19833-5_34},
  url          = {https://arxiv.org/abs/2207.01696}
}

@inproceedings{t2mgpt2023,
  title     = {Generating Human Motion from Textual Descriptions with Discrete Representations},
  author    = {Zhang, Jianrong and Zhang, Yangsong and Cun, Xiaodong and Zhang, Yong and Zhao, Hongwei and Lu, Hongtao and Shen, Xi and Shan, Ying},
  booktitle = {Proceedings of the IEEE/CVF Conference on Computer Vision and Pattern Recognition (CVPR)},
  pages     = {14730--14740},
  year      = {2023},
  doi       = {10.1109/CVPR52729.2023.01415},
  url       = {https://openaccess.thecvf.com/content/CVPR2023/html/Zhang_Generating_Human_Motion_From_Textual_Descriptions_With_Discrete_Representations_CVPR_2023_paper.html}
}

@InProceedings{mindeye2,
  title     = {{M}ind{E}ye2: Shared-Subject Models Enable f{MRI}-To-Image With 1 Hour of Data},
  author    = {Scotti, Paul Steven and Tripathy, Mihir and Torrico, Cesar and Kneeland, Reese and Chen, Tong and Narang, Ashutosh and Santhirasegaran, Charan and Xu, Jonathan and Naselaris, Thomas and Norman, Kenneth A. and Abraham, Tanishq Mathew},
  booktitle = {Proceedings of the 41st International Conference on Machine Learning},
  pages     = {44038--44059},
  year      = {2024},
  editor    = {Salakhutdinov, Ruslan and Kolter, Zico and Heller, Katherine and Weller, Adrian and Oliver, Nuria and Scarlett, Jonathan and Berkenkamp, Felix},
  volume    = {235},
  series    = {Proceedings of Machine Learning Research},
  month     = {21--27 Jul},
  publisher = {PMLR},
  url       = {https://proceedings.mlr.press/v235/scotti24a.html}
}

@inproceedings{e2t,
  title     = "Enhancing {EEG}-to-Text Decoding through Transferable Representations from Pre-trained Contrastive {EEG}-Text Masked Autoencoder",
  author    = "Wang, Jiaqi  and
               Song, Zhenxi  and
               Ma, Zhengyu  and
               Qiu, Xipeng  and
               Zhang, Min  and
               Zhang, Zhiguo",
  editor    = "Ku, Lun-Wei  and
               Martins, Andre  and
               Srikumar, Vivek",
  booktitle = "Proceedings of the 62nd Annual Meeting of the Association for Computational Linguistics (Volume 1: Long Papers)",
  month     = aug,
  year      = "2024",
  address   = "Bangkok, Thailand",
  publisher = "Association for Computational Linguistics",
  url       = "https://aclanthology.org/2024.acl-long.393/",
  doi       = "10.18653/v1/2024.acl-long.393",
  pages     = "7278--7292"
}

@article{gwin2011electrocortical,
  title        = {Electrocortical activity is coupled to gait cycle phase during treadmill walking},
  author       = {Gwin, Joseph T. and Gramann, Klaus and Makeig, Scott and Ferris, Daniel P.},
  journal      = {NeuroImage},
  volume       = {54},
  number       = {2},
  pages        = {1289--1296},
  year         = {2011},
  doi          = {10.1016/j.neuroimage.2010.08.066}
}

@article{knaepen2015grcp,
  title   = {Temporal and spatial organization of gait-related electrocortical potentials},
  author  = {Knaepen, Kristel and Mierau, Andreas and Fernandez Tellez, Helio and Lefeber, Dirk and Meeusen, Romain},
  journal = {Neuroscience Letters},
  volume  = {599},
  pages   = {75--80},
  year    = {2015},
  doi     = {10.1016/j.neulet.2015.05.036},
  url     = {https://doi.org/10.1016/j.neulet.2015.05.036}
}

@article{duan2024e2h,
  title         = {{E2H}: A Two-Stage Non-Invasive Neural Signal Driven Humanoid Robotic Whole-Body Control Framework},
  author        = {Duan, Yiqun and Zhang, Qiang and Zhou, Jinzhao and Sun, Jingkai and Jiang, Xiaowei and Cao, Jiahang and Wang, Jiaxu and Yang, Yiqian and Zhao, Wen and Han, Gang and Guo, Yijie and Lin, Chin-Teng},
  journal       = {arXiv preprint arXiv:2410.02141},
  year          = {2024},
  eprint        = {2410.02141},
  archivePrefix = {arXiv},
  primaryClass  = {cs.RO},
  url           = {https://arxiv.org/abs/2410.02141}
}

@article{kriegeskorte2008rsa,
  title        = {Representational similarity analysis---connecting the branches of systems neuroscience},
  author       = {Kriegeskorte, Nikolaus and Mur, Marieke and Bandettini, Peter A.},
  journal      = {Frontiers in Systems Neuroscience},
  volume       = {2},
  pages        = {4},
  year         = {2008},
  doi          = {10.3389/neuro.06.004.2008}
}

@article{robinson2016noninvasive_smc_mag,
  title   = {Noninvasive Brain--Computer Interface: Decoding Arm Movement Kinematics and Motor Control},
  author  = {Robinson, Neethu and Vinod, A. P.},
  journal = {IEEE Systems, Man, and Cybernetics Magazine},
  volume  = {2},
  number  = {4},
  pages   = {4--16},
  year    = {2016},
  month   = oct,
  doi     = {10.1109/MSMC.2016.2576638},
  url     = {https://www.ieeesmc.org/wp-content/uploads/2014/03/SMC-Magazine-BCI-Article-October-2016.pdf}
}

@article{schirrmeister2017deep,
  title        = {Deep learning with convolutional neural networks for {EEG} decoding and visualization},
  author       = {Schirrmeister, Robin Tibor and Springenberg, Jost Tobias and Fiederer, Lukas D. J. and Glasstetter, Martin and Eggensperger, Katharina and Tangermann, Michael and Hutter, Frank and Burgard, Wolfram and Ball, Tonio},
  journal      = {Human Brain Mapping},
  volume       = {38},
  number       = {11},
  pages        = {5391--5420},
  year         = {2017},
  doi          = {10.1002/hbm.23730}
}

@article{lawhern2018eegnet,
  title        = {{EEGNet}: A compact convolutional neural network for {EEG}-based brain--computer interfaces},
  author       = {Lawhern, Vernon J. and Solon, Amelia J. and Waytowich, Nicholas R. and Gordon, Stephen M. and Hung, Chou P. and Lance, Brent J.},
  journal      = {Journal of Neural Engineering},
  volume       = {15},
  number       = {5},
  pages        = {056013},
  year         = {2018},
  doi          = {10.1088/1741-2552/aace8c}
}

@article{song2023eegconformer,
  title        = {{EEG} Conformer: Convolutional Transformer for {EEG} Decoding and Visualization},
  author       = {Song, Yonghao and Zheng, Qingqing and Liu, Bingchuan and Gao, Xiaorong},
  journal      = {IEEE Transactions on Neural Systems and Rehabilitation Engineering},
  volume       = {31},
  pages        = {710--719},
  year         = {2023},
  doi          = {10.1109/TNSRE.2022.3230250}
}

@article{wang2025eeg2gait,
  title        = {{EEG2GAIT}: A Hierarchical Graph Convolutional Network for {EEG}-based Gait Decoding},
  author       = {Fu, Xi and Liu, Rui and Wai, Aung Aung Phyo and Pulferer, Hannah and Robinson, Neethu and M{\"u}ller-Putz, Gernot R. and Guan, Cuntai},
  journal      = {IEEE Transactions on Neural Systems and Rehabilitation Engineering},
  volume       = {34},
  pages        = {441--454},
  year         = {2026},
  doi          = {10.1109/TNSRE.2025.3647101},
  url          = {https://arxiv.org/abs/2504.03757}
}

@article{tortora2020lstmwalk,
  title        = {Deep learning-based {BCI} for gait decoding from {EEG} with {LSTM} recurrent neural network},
  author       = {Tortora, Stefano and Ghidoni, Stefano and Chisari, Carmelo and Micera, Silvestro and Artoni, Fiorenzo},
  journal      = {Journal of Neural Engineering},
  volume       = {17},
  number       = {4},
  pages        = {046011},
  year         = {2020},
  month        = jul,
  publisher    = {IOP Publishing},
  doi          = {10.1088/1741-2552/ab9842},
  url          = {https://doi.org/10.1088/1741-2552/ab9842}
}

@article{fu2025neurodygait,
  title         = {Zero-Shot {EEG}-to-Gait Decoding via Phase-Aware Representation Learning},
  author        = {Fu, Xi and Jiang, Weibang and Liu, Rui and M{\"u}ller-Putz, Gernot R. and Guan, Cuntai},
  journal       = {arXiv preprint arXiv:2506.22488v1},
  year          = {2025},
  eprint        = {2506.22488v1},
  archivePrefix = {arXiv},
  primaryClass  = {eess.SP},
  url           = {https://arxiv.org/abs/2506.22488v1}
}

@article{zhang2024mtrt,
  title        = {{MTRT}: Motion Trajectory Reconstruction Transformer for {EEG}-Based {BCI} Decoding},
  author       = {Wang, Pengpai and Li, Zhongnian and Gong, Peiliang and Zhou, Yueying and Chen, Fang and Zhang, Daoqiang},
  journal      = {IEEE Transactions on Neural Systems and Rehabilitation Engineering},
  volume       = {31},
  pages        = {2349--2358},
  year         = {2023},
  doi          = {10.1109/TNSRE.2023.3275172}
}

@inproceedings{jiang2025neurolm,
  title     = {{NeuroLM}: A Universal Multi-task Foundation Model for Bridging the Gap between Language and {EEG} Signals},
  author    = {Jiang, Wei-Bang and Wang, Yansen and Lu, Bao-Liang and Li, Dongsheng},
  booktitle = {International Conference on Learning Representations (ICLR)},
  year      = {2025},
  url       = {https://arxiv.org/abs/2409.00101}
}

@article{farshchian2022nomad,
  title        = {Stabilizing brain-computer interfaces through alignment of latent dynamics},
  author       = {Karpowicz, Brianna M. and Ali, Yahia H. and Wimalasena, Lahiru N. and Sedler, Andrew R. and Keshtkaran, Mohammad Reza and Bodkin, Kevin and Ma, Xuan and Rubin, Daniel B. and Williams, Ziv M. and Cash, Sydney S. and Hochberg, Leigh R. and Miller, Lee E. and Pandarinath, Chethan},
  journal      = {Nature Communications},
  volume       = {16},
  number       = {1},
  pages        = {4662},
  year         = {2025},
  doi          = {10.1038/s41467-025-59652-y}
}

@inproceedings{vaswani2017attention,
  title     = {Attention Is All You Need},
  author    = {Vaswani, Ashish and Shazeer, Noam and Parmar, Niki and Uszkoreit, Jakob and Jones, Llion and Gomez, Aidan N. and Kaiser, {\L}ukasz and Polosukhin, Illia},
  booktitle = {Advances in Neural Information Processing Systems (NeurIPS)},
  volume    = {30},
  pages     = {5998--6008},
  year      = {2017},
  url       = {https://arxiv.org/abs/1706.03762}
}

@article{delorme2004eeglab,
  title        = {{EEGLAB}: an open source toolbox for analysis of single-trial {EEG} dynamics including independent component analysis},
  author       = {Delorme, Arnaud and Makeig, Scott},
  journal      = {Journal of Neuroscience Methods},
  volume       = {134},
  number       = {1},
  pages        = {9--21},
  year         = {2004},
  doi          = {10.1016/j.jneumeth.2003.10.009}
}

@inproceedings{guo2022diversehumanmotion,
  title        = {Generating Diverse and Natural {3D} Human Motions from Text},
  author       = {Guo, Chuan and Zou, Shihao and Zuo, Xinxin and Wang, Sen and Ji, Wei and Li, Xingyu and Cheng, Li},
  booktitle    = {Proceedings of the IEEE/CVF Conference on Computer Vision and Pattern Recognition (CVPR)},
  pages        = {5152--5161},
  year         = {2022},
  doi          = {10.1109/CVPR52688.2022.00509},
  url          = {https://openaccess.thecvf.com/content/CVPR2022/html/Guo_Generating_Diverse_and_Natural_3D_Human_Motions_From_Text_CVPR_2022_paper.html}
}

@article{banville2023contrawr,
  title   = {Self-supervised electroencephalogram representation learning for automatic sleep staging: model development and evaluation study},
  author  = {Yang, Chaoqi and Xiao, Cao and Westover, M. Brandon and Sun, Jimeng},
  journal = {JMIR AI},
  volume  = {2},
  number  = {1},
  pages   = {e46769},
  year    = {2023},
  doi     = {10.2196/46769},
  url     = {https://ai.jmir.org/2023/1/e46769}
}

@inproceedings{zhu2025motiongpt3,
  title         = {{MotionGPT3}: Human Motion as a Second Modality},
  author        = {Zhu, Bingfan and Jiang, Biao and Wang, Sunyi and Tang, Shixiang and Chen, Tao and Luo, Linjie and Zheng, Youyi and Chen, Xin},
  booktitle     = {International Conference on Learning Representations (ICLR)},
  year          = {2026},
  eprint        = {2506.24086},
  archivePrefix = {arXiv},
  primaryClass  = {cs.CV},
  url           = {https://proceedings.iclr.cc/paper_files/paper/2026/hash/1551c01d7a3d0bf21e2518331e9f7074-Abstract-Conference.html}
}

@article{he2018treadmill,
  title   = {A mobile brain-body imaging dataset recorded during treadmill walking with a brain-computer interface},
  author  = {He, Yongtian and Luu, Trieu Phat and Nathan, Kevin and Nakagome, Sho and Contreras-Vidal, Jose L.},
  journal = {Scientific Data},
  volume  = {5},
  pages   = {180074},
  year    = {2018},
  doi     = {10.1038/sdata.2018.74}
}

@article{thapa2025freewill,
  title   = {A large electroencephalogram database of freewill reaching and grasping tasks for brain machine interfaces},
  author  = {Thapa, Bhoj Raj and Boggess, John and Bae, Jihye},
  journal = {Scientific Data},
  volume  = {12},
  pages   = {1760},
  year    = {2025},
  doi     = {10.1038/s41597-025-06039-9}
}

@inproceedings{luu2016unscented,
  title     = {Unscented Kalman filter for neural decoding of human treadmill walking from non-invasive electroencephalography},
  author    = {Luu, Trieu Phat and He, Yongtian and Nakagame, Sho and Gorges, Jeffrey and Nathan, Kevin and Contreras-Vidal, Jose L.},
  booktitle = {2016 38th Annual International Conference of the IEEE Engineering in Medicine and Biology Society (EMBC)},
  pages     = {1548--1551},
  year      = {2016},
  publisher = {IEEE},
  doi       = {10.1109/EMBC.2016.7591006}
}

@inproceedings{wang2025vqvla,
  title         = {{VQ-VLA}: Improving Vision-Language-Action Models via Scaling Vector-Quantized Action Tokenizers},
  author        = {Wang, Yating and Zhu, Haoyi and Liu, Mingyu and Yang, Jiange and Fang, Hao-Shu and He, Tong},
  booktitle     = {Proceedings of the IEEE/CVF International Conference on Computer Vision (ICCV)},
  pages         = {11089--11099},
  year          = {2025},
  doi           = {10.1109/ICCV51701.2025.01032},
  eprint        = {2507.01016},
  archivePrefix = {arXiv},
  primaryClass  = {cs.RO},
  url           = {https://openaccess.thecvf.com/content/ICCV2025/html/Wang_VQ-VLA_Improving_Vision-Language-Action_Models_via_Scaling_Vector-Quantized_Action_Tokenizers_ICCV_2025_paper.html}
}

@article{nakagome2020comparison,
  title = {An empirical comparison of neural networks and machine learning algorithms for EEG gait decoding},
  author = {Nakagome, Sho and Luu, Trieu Phat and He, Yongtian and Ravindran, Akshay Sujatha and Contreras-Vidal, Jose L.},
  journal = {Scientific Reports},
  volume = {10},
  pages = {4372},
  year = {2020},
  doi = {10.1038/s41598-020-60932-4},
  url = {https://www.nature.com/articles/s41598-020-60932-4}
}

@article{li2009ukf,
  title = {Unscented Kalman Filter for Brain-Machine Interfaces},
  author = {Li, Zheng and O'Doherty, Joseph E. and Hanson, Timothy L. and Lebedev, Mikhail A. and Henriquez, Craig S. and Nicolelis, Miguel A. L.},
  journal = {PLOS ONE},
  volume = {4},
  number = {7},
  pages = {e6243},
  year = {2009},
  doi = {10.1371/journal.pone.0006243},
  url = {https://doi.org/10.1371/journal.pone.0006243}
}

@article{song2021sttransformer,
  title = {Transformer-based Spatial-Temporal Feature Learning for {EEG} Decoding},
  author = {Song, Yonghao and Jia, Xueyu and Yang, Lie and Xie, Longhan},
  journal = {arXiv preprint arXiv:2106.11170},
  year = {2021},
  eprint = {2106.11170},
  archivePrefix = {arXiv},
  primaryClass = {eess.SP},
  url = {https://arxiv.org/abs/2106.11170}
}

@article{peng2026eeg2motion,
  title = {{EEG2MOTION}: Towards Open-Vocabulary Human Motion Synthesis from Non-invasive Brain Signals},
  author = {Peng, Yulong and Pan, Yijian and Yang, Yuqi and Zheng, Nenggan and Chen, Weidong and Hu, Xiaoling and Zhang, Shaomin},
  journal = {arXiv preprint arXiv:2608.14754},
  year = {2026},
  url = {https://arxiv.org/abs/2608.14754}
}
\endgroup



\clearpage
\appendix
\section*{\textbf{Appendix}}

\section{Related Work}

\subsection{EEG-based motion decoding}

Decoding movement-related information from non-invasive EEG has long been studied in BCI research. Early studies showed that low-dimensional kinematic variables, such as movement direction, hand velocity, or joint trajectories, can be decoded from EEG in controlled paradigms~\citep{bradberry2010reconstructing,robinson2016noninvasive_smc_mag}. Deep learning further enabled end-to-end spatiotemporal modeling of minimally processed EEG using CNN, RNN, and compact convolutional architectures~\citep{schirrmeister2017deep,lawhern2018eegnet}, while recent Transformer-based models improve long-range temporal modeling for EEG decoding~\citep{song2023eegconformer}.

EEG-based motor decoding has been studied across a range of movement paradigms. Upper-limb decoding commonly focuses on hand movement, reaching, or low-dimensional arm trajectories~\citep{bradberry2010reconstructing,robinson2016noninvasive_smc_mag}, with sequence models such as MTRT reconstructing continuous arm motion using Transformer-based temporal modeling~\citep{zhang2024mtrt}. Lower-limb and gait decoding methods study cyclic locomotion dynamics, motion artifacts, and nonstationarity, using CNN/LSTM architectures~\citep{tortora2020lstmwalk}, hierarchical or graph-based designs~\citep{wang2025eeg2gait}, and phase-aware representation learning for cross-session or cross-subject robustness~\citep{fu2025neurodygait}. Recent work has also explored structured interfaces for whole-body or humanoid-oriented control, such as E2H, which uses an EEG-to-keyword and keyword-to-motion pipeline~\citep{duan2024e2h}. Together, these studies provide important foundations for modeling neural activity associated with upper-limb, lower-limb, gait, and whole-body movement.

For the gait comparison, NeuroDyGait serves as the later task-specific reference. Our comparison covers direct regression and learned continuous intermediate representations; it does not establish superiority over every possible adapted generative model.

\subsection{Motion and action tokenization}

Discrete representations have become an effective tool for modeling continuous motion and action sequences. In text-to-motion generation, vector-quantized autoencoders tokenize high-dimensional human motion into compact codebooks~\citep{oord2017vqvae}, enabling Transformer-based autoregressive generation conditioned on textual or semantic inputs~\citep{guo2022tm2t,t2mgpt2023}. These methods show that learned motion codebooks can absorb low-level kinematic details and provide a structured latent space for long-horizon motion synthesis. \rev{EEG2MOTION~\citep{peng2026eeg2motion} extends this text-to-motion paradigm to motion synthesis from EEG recorded during action observation. It combines a pretrained HumanML3D VQ-based motion prior with global EEG--motion contrastive alignment and EEG-conditioned masked-token generation.}

Motion and action tokenization has also been explored in vision-language-action and robotic control models. VQ-style action quantization, as in VQ-VLA ~\citep{wang2025vqvla}, discretizes low-dimensional control trajectories into reusable action tokens. This suggests that quantization can serve as a general mechanism for representing both high-dimensional human motion and lower-dimensional robotic action sequences.

\subsection{Representation alignment for neural decoding}

Robust neural decoding often benefits from learning structured latent representations rather than relying solely on direct output supervision. Recent EEG foundation models discretize EEG into neural tokens and train sequence models with language-model-like objectives, enabling unified inference across multiple EEG tasks~\citep{jiang2025neurolm}. Related studies show that aligning latent neural dynamics across sessions can improve decoding stability and robustness~\citep{farshchian2022nomad}. In brain-to-image reconstruction, approaches such as MindEye demonstrate that contrastive alignment between brain activity and semantic embedding spaces can learn informative neural representations for retrieval and reconstruction~\citep{mindeye2}. Similar alignment strategies have also been explored in EEG-to-text decoding to bridge neural activity and language representations~\citep{e2t}.

NeuroMotion learns discrete primitives from motion and aligns continuous EEG representations to their codebook vectors. These primitives provide both representation-level supervision and intermediate prediction targets for continuous motion reconstruction. EEG tokenizers used for pretraining or classification therefore address a different intermediate representation.Likewise, generic motion generation and EEG foundation encoders require an EEG-conditioning or synchronized kinematic decoding interface before they can be evaluated on this task. Testing such adaptations is useful future work; their absence does not imply that they are ineffective.

\section{Experimental Settings}
\label{app:settings}
\subsection{Dataset Acquisition and Session Protocols}
\label{app:dataset_protocols}

FBM records EEG and full-body kinematics during continuous overground multi-terrain locomotion, including level walking, ramps, and stairs~\citep{zhu2018fbm}. We use subjects \texttt{s02}--\texttt{s10}, excluding \texttt{s01} due to synchronization issues. For each subject, the last available session is held out for testing. The dataset does not provide explicit visit/day metadata sufficient to establish a cross-day split, so we interpret this protocol as session-disjoint cross-trial evaluation rather than verified cross-day generalization.

TM records 60-channel EEG and bilateral lower-limb joint angles during treadmill walking and closed-loop BCI control~\citep{he2018treadmill}. We evaluate six subjects, excluding \texttt{S03} and \texttt{S04} due to data quality issues. Sessions 01 and 02 were recorded during the first visit, while Session 03 was recorded during a second visit within seven days. Training on Session 01/02 and testing on Session 03 therefore provides a cross-day evaluation.

FR records 31-channel EEG and three-axis wrist acceleration during voluntary reaching-and-grasping~\citep{thapa2025freewill}. We evaluate \texttt{s05}, \texttt{s07}, \texttt{s13}, \texttt{s15}, and \texttt{s22}, the five subjects with complete three-session recordings. Sessions 1--3 were recorded on different days; training on the earlier sessions and testing on Session 3 therefore constitutes cross-day evaluation.

Accordingly, the main TM and FR protocols evaluate session-disjoint cross-day generalization, whereas FBM provides session-disjoint evaluation without verified day separation. None of these protocols establishes stability over multi-week or multi-month intervals.

\subsection{Model and Training Details}
\label{app:implementation}

\paragraph{Signal preprocessing.}
For all datasets, kinematic signals are low-pass filtered at 5\,Hz, downsampled to 20\,Hz, and temporally aligned with EEG. FBM motion is converted into the 263-D HumanML3D-style representation described in Sec.~\ref{sec:motion_repr_263}, while TM and FR motion targets are min-max normalized to $[0,1]$. EEG signals are band-pass filtered at 0.1--48\,Hz, re-referenced using common average reference (CAR)~\citep{delorme2004eeglab}, and downsampled to 100\,Hz.Channel-wise z-score normalization is then applied as the first operation of the EEG encoder rather than as an offline preprocessing step.

\paragraph{Windowing and token length.}
All datasets are processed into temporally aligned EEG-motion windows of length 2\,s with a stride of 0.5\,s. Kinematic signals are downsampled to 20\,Hz, so each motion window contains 40 frames. The motion tokenizer temporally downsamples each window into a sequence of motion tokens. The default token counts are $N_e=N_m=10$ for FBM/TM and $N_e=N_m=5$ for FR, corresponding to nominal durations of 0.2\,s and 0.4\,s per token. The model predicts one motion-token sequence per EEG window.

\paragraph{Architecture hyperparameters.}
For FBM, we use a VQ codebook size of 32. The EEG encoder hidden dimension and E2M hidden dimension are both set to 512. The EEG encoder contains 6 Transformer layers, and the E2M Transformer contains 16 layers with 8 attention heads.

For TM and FR, which contain lower-dimensional kinematic targets, we use smaller codebooks and model widths. Specifically, TM uses a VQ codebook size of 16, while FR uses a VQ codebook size of 12. For both datasets, the EEG encoder and E2M hidden dimensions are set to 256. The EEG encoder contains 6 Transformer layers, and the E2M Transformer contains 16 layers with 8 attention heads.

\paragraph{Stage 1: VQ motion prior training.}
We train a separate motion tokenizer for each dataset using only the kinematic training data. Test sessions used in EEG-to-motion evaluation are excluded from VQ training. The VQ tokenizer is trained with AdamW using learning rate $2\times10^{-4}$, batch size 1024, weight decay $1\times10^{-3}$, 100k iterations, 1k warmup iterations, and a cosine learning-rate schedule. The reconstruction loss weight is set to 1.0, and the commitment loss weight is set to 0.02. After training, the VQ encoder, codebook, and decoder define the dataset-specific discrete motion prior.

\paragraph{Stage 2: EEG-to-motion token learning.}
With the motion tokenizer frozen, we train the EEG encoder, neuro-motion alignment module, and E2M Transformer to predict motion-token sequences from EEG. The training objective is
\begin{equation}
\mathcal{L}_{\mathrm{stage2}}
=
\mathcal{L}_{\mathrm{tok}}
+
\lambda_{\mathrm{pcl}}\mathcal{L}_{\mathrm{pcl}}
+
\lambda_{\mathrm{supcon}}\mathcal{L}_{\mathrm{supcon}}.
\end{equation}
We set $\lambda_{\mathrm{pcl}}=1.0$ and $\lambda_{\mathrm{supcon}}=1.0$, and use temperature $\tau=0.07$ for both contrastive objectives. Stage 2 is trained with AdamW using learning rate $1\times10^{-3}$, batch size 512, weight decay $1\times10^{-3}$, 12k iterations, 1k warmup iterations, and a cosine learning-rate schedule.

\paragraph{Stage 3: geometric refinement.}
Starting from the Stage 2 checkpoint, we perform geometric refinement by optimizing the decoded continuous motion with an MSE loss. During this stage, the EEG encoder, E2M Transformer, and VQ decoder are updated, while the VQ encoder and codebook remain fixed. We use AdamW with learning rate $1\times10^{-5}$, batch size 64, weight decay $1\times10^{-3}$, 2k iterations, 1k warmup iterations, and a cosine learning-rate schedule.

For FBM, Stage-3 MSE supervises only the 30-D end-effector subset in Appendix~\ref{app:metrics}; TM/FR supervise all 6/3 target dimensions. The FBM 263-D supervision variant in Appendix~\ref{app:refinement_scope} changes only this supervision scope.

\paragraph{Training cost.}
All experiments are trained on a single NVIDIA L4 GPU. Stage 1 is run once per dataset and takes approximately 6\,h for FBM and 2\,h for TM/FR. Stages 2--3 are subject-specific. With subject-dependent convergence and early stopping, Stage 2 runs for 2,000--12,000 iterations (10--60\,min), and Stage 3 for 500--2,000 iterations (3--12\,min), within the nominal budgets above. Subject-specific training therefore takes 13--72\,min. Original and additional experiments share the same preprocessing, split, normalization, and training settings except for explicitly controlled factors.

\subsection{Motion representation (263-D)}
\label{sec:motion_repr_263}

Following prior tokenized motion generation pipelines~\citep{t2mgpt2023}, each FBM motion frame uses the standard 263-D HumanML3D-style representation:
\begin{equation}
\mathbf{x}=\big(\dot r_a,\dot r_x,\dot r_z,r_y,
\mathbf{p}_{\neg r},\mathbf{R}_{\neg r},\mathbf{v},\mathbf{c}_f\big)
\in\mathbb{R}^{263}.
\end{equation}
For $j=22$ joints, the root block contains yaw angular velocity (1), planar root velocity expressed in the root frame (2), and root height (1). Root-relative positions $\mathbf{p}_{\neg r}\in\mathbb{R}^{3(j-1)}$ and continuous 6D rotations $\mathbf{R}_{\neg r}\in\mathbb{R}^{6(j-1)}$ exclude the root joint. Local joint velocities $\mathbf{v}\in\mathbb{R}^{3j}$ include all joints, and $\mathbf{c}_f\in\mathbb{R}^{4}$ contains foot-contact indicators. The concatenation order is root features, non-root positions, non-root rotations, all-joint velocities, and contacts. Consequently,
\[
1+2+1+3(j-1)+6(j-1)+3j+4
=4+63+126+66+4=263.
\]
\subsection{Baseline Adaptation and Training}
\label{app:baseline_adaptation}

For a controlled comparison, all baselines use the same EEG preprocessing, session-wise train/test splits, dataset-specific target representations, normalization, and evaluation metrics as NeuroMotion. Each baseline retains its original backbone, while only the dataset-specific input interface and regression output head are adapted where necessary.

On FBM, all methods predict the same 263-D HumanML3D-style representation over 40 motion frames. For task-specific EEG-to-motion baselines, namely NeuroDyGait and MTRT, we retain their continuous sequence-decoding structures, adapting the input and output channel dimensions and making moderate adjustments to internal network depth. For the general-purpose EEG baselines (DeepConvNet, EEGConformer, STTransformer, and ContraWR), we use a common MLP readout on FBM with widths $256\rightarrow2048\rightarrow10{,}520$, followed by reshaping to $40\times263$. On FR and TM, these general-purpose baselines instead use a single linear projection from their final feature representations to the complete motion window, followed by reshaping to $40\times3$ and $40\times6$, respectively. The corresponding targets are the normalized 3-D wrist-acceleration sequence on FR and 6-D lower-limb joint-angle sequence on TM.

For each \rev{deep-learning} baseline, dropout is selected from $\{0.0,0.1,\ldots,0.9\}$. Training uses learning rate $10^{-4}$, cosine scheduling, a 200-epoch budget, and early-stopping patience 10. These controls standardize the prediction task and baseline optimization protocol; backbone capacities and the staged NeuroMotion training procedure are not forced to be identical across methods.

\rev{The added LR, RR, and UKF baselines are adapted from the public EEG-neural-decoding implementation~\citep{nakagome2020comparison}. LR/RR flatten the complete 2-s EEG window and jointly predict all 40 motion frames; RR uses $\alpha=1$. UKF preserves frame-wise recursive state estimation and the nonlinear observation formulation of \citet{li2009ukf}. Reformulating UKF as joint whole-window regression would substantially change its state and observation models and recursive inference procedure, so we retain the reference decoding formulation. UKF uses a first-order state model, four EEG taps at 20\,Hz, a one-frame EEG lag, and transition/observation ridge coefficients of 1 and 5. Each window starts from the training-state mean. EEG and target standardization are fitted on the training split. All three methods use the same dataset-specific targets, session splits, and evaluation metrics as the other methods.}

Regarding EEG2GAIT, NeuroDyGait directly compares against it, reports stronger gait-decoding performance, and shares the same first author; we therefore use NeuroDyGait as the later and stronger gait-specific reference. Moreover, the EEG2GAIT release lacks components required for full reproduction without undocumented assumptions.

\rev{EEG2MOTION~\citep{peng2026eeg2motion} is not included as a directly adapted baseline because its pretrained motion generator is coupled to the HumanML3D motion representation and codebook. It cannot be directly reused for TM joint angles or FR wrist accelerations, which require task-specific motion tokenizers, decoders, and corresponding generator adaptation. FBM also requires adaptation to the recorded skeleton and motion domain. In addition, EEG2MOTION synthesizes observed actions from action-observation EEG and emphasizes semantic matching and generative quality, evaluated using R-Precision, MM-Dist, FID, and Diversity. Our task reconstructs synchronized trajectories from EEG recorded during actual movement, with PCC, $R^2$, and RMSE as the primary metrics.}

\section{Evaluation Metrics}
\label{app:metrics}

We evaluate EEG-to-motion decoding with RMSE, PCC, and $R^2$ on all three datasets. For FBM, these reconstruction metrics are computed on the 30-D end-effector position subset of the 263-D HumanML3D-style motion vector, corresponding to the XYZ coordinates of 10 limb joints: LKnee, RKnee, LHeel, RHeel, LFoot, RFoot, LElbow, RElbow, LWrist, and RWrist. The 0-based indices are
\[
\mathcal{D}_{\mathrm{FBM}}=
[13{:}19,\;22{:}28,\;31{:}37,\;55{:}67],
\]
where each range is a zero-based, half-open slice, giving 30 dimensions in total. For TM, the metrics are computed on the 6-D bilateral lower-limb joint-angle sequence from hip, knee, and ankle goniometers. For FR, the metrics are computed on the 3-D wrist-acceleration sequence. For FBM, we additionally report FID and Diversity in a pretrained motion-embedding space. We compute each metric per subject and then average across subjects; reported standard deviations describe between-subject variability. Paired comparisons use two-sided paired $t$-tests across subjects, with uncorrected $p$-values reported in Appendix~\ref{app:cohort_statistics}.

\subsection{Root Mean Squared Error (RMSE)}

Given predicted motion $\hat{\mathbf{Y}}\in\mathbb{R}^{B\times T\times D}$ and ground-truth motion $\mathbf{Y}\in\mathbb{R}^{B\times T\times D}$, where $D=30$ for FBM reconstruction metrics, $D=6$ for TM, and $D=3$ for FR, we first flatten the batch and temporal axes into $n=1,\ldots,BT$. RMSE is computed independently for each evaluated dimension and then averaged:
\begin{equation}
\mathrm{RMSE}^{(d)}
=
\sqrt{
\frac{1}{BT}
\sum_{n=1}^{BT}
\left(
\hat{y}^{(d)}_n-y^{(d)}_n
\right)^2
},
\qquad
\mathrm{RMSE}
=
\frac{1}{D}
\sum_{d=1}^{D}
\mathrm{RMSE}^{(d)}.
\end{equation}
Lower RMSE indicates more accurate continuous motion reconstruction.

\subsection{Pearson Correlation Coefficient (PCC)}

PCC measures temporal correlation between predicted and ground-truth motion signals. We compute PCC separately for each evaluated motion dimension and report the average across dimensions:
\begin{equation}
\mathrm{PCC}^{(d)}
=
\frac{
\sum_{n}
(\hat{y}^{(d)}_n-\bar{\hat{y}}^{(d)})
(y^{(d)}_n-\bar{y}^{(d)})
}{
\sqrt{\sum_{n}(\hat{y}^{(d)}_n-\bar{\hat{y}}^{(d)})^2}
\sqrt{\sum_{n}(y^{(d)}_n-\bar{y}^{(d)})^2}
},
\qquad
\mathrm{PCC}
=
\frac{1}{D}
\sum_{d=1}^{D}
\mathrm{PCC}^{(d)}.
\end{equation}
Here, $n$ indexes all sample-frame pairs. Higher PCC impies better temporal agreement with ground truth.

\subsection{Coefficient of Determination ($R^2$)}

$R^2$ is also computed independently for each evaluated motion dimension and then averaged:
\begin{equation}
R^{2,(d)}
=
1-
\frac{
\sum_{n}
\left(
y^{(d)}_n-\hat{y}^{(d)}_n
\right)^2
}{
\sum_{n}
\left(
y^{(d)}_n-\bar{y}^{(d)}
\right)^2
},
\qquad
R^2
=
\frac{1}{D}
\sum_{d=1}^{D}
R^{2,(d)}.
\end{equation}
Higher $R^2$ indicates better explained variance, and negative values indicate performance worse than predicting the mean for that dimension.

\subsection{Fr\'echet Inception Distance (FID) for FBM}

For FBM, we evaluate distribution-level full-body motion quality using FID in a fixed pretrained motion-embedding space. Specifically, we use the motion encoder from the evaluator released by ~\citet{guo2022diversehumanmotion}, which is commonly used in text-to-motion evaluation. Unlike RMSE, PCC, and $R^2$, FID is computed on the full 263-D HumanML3D-style motion sequence after evaluator preprocessing. Given generated motion embeddings $\{\hat{\mathbf{z}}_n\}_{n=1}^{N}$ and real motion embeddings $\{\mathbf{z}_n\}_{n=1}^{N}$, we compute empirical means and covariances $(\boldsymbol{\mu}_g,\boldsymbol{\Sigma}_g)$ and $(\boldsymbol{\mu}_r,\boldsymbol{\Sigma}_r)$. FID is defined as:
\begin{equation}
\mathrm{FID}
=
\left\lVert
\boldsymbol{\mu}_g-\boldsymbol{\mu}_r
\right\rVert_2^2
+
\mathrm{Tr}
\left(
\boldsymbol{\Sigma}_g
+
\boldsymbol{\Sigma}_r
-
2
(\boldsymbol{\Sigma}_g\boldsymbol{\Sigma}_r)^{1/2}
\right).
\end{equation}
Lower FID indicates closer distributions in this evaluator space. Because the evaluator was trained on a different skeleton/data domain from FBM, it is a complementary measure and should not be interpreted as direct evidence of paired trajectory accuracy; small differences with large subject variability warrant caution.

\subsection{Diversity for FBM}

For FBM, we report Diversity to measure the variability of generated full-body motions in the same Guo et al. motion-encoder embedding space~\citep{guo2022diversehumanmotion}. Diversity is computed on full 263-D HumanML3D-style motion sequences after evaluator preprocessing. Given generated embeddings $\{\hat{\mathbf{z}}_n\}_{n=1}^{N}$, Diversity is computed as:
\begin{equation}
\mathrm{Diversity}
=
\frac{2}{N(N-1)}
\sum_{1\le i<j\le N}
\left\lVert
\hat{\mathbf{z}}_i-\hat{\mathbf{z}}_j
\right\rVert_2.
\end{equation}
We also report the Diversity of ground-truth FBM motions as a reference. A generated set with Diversity closer to the real-motion reference indicates better coverage of the natural motion distribution.

\section{Additional Experimental Results}

\subsection{Extended Ablation Studies}
\label{app:extended_ablation}

\begin{table}[tbp]
\centering
\caption{Codebook capacity on three subjects per dataset: FBM (S02/S07/S10), TM (S02/S06/S07), and FR (S05/S07/S22). Entries are mean $\pm$ between-subject standard deviation. The default sizes are $K=32/16/12$ for FBM/TM/FR.}
\label{tab:vq_codebook_ablation}

\small
\setlength{\tabcolsep}{5pt}
\renewcommand{\arraystretch}{1.05}

\begin{tabular}{@{}llccc@{}}
\toprule
Dataset & $K$ & PCC$\uparrow$ & $R^2\uparrow$ & RMSE$\downarrow$ \\
\midrule

FBM & 12
& \mstd{0.449}{0.382}
& \mstd{0.120}{0.445}
& \mstd{0.785}{0.354} \\

FBM & 16
& \mstd{0.454}{0.403}
& \mstd{0.135}{0.470}
& \mstd{0.771}{0.354} \\

FBM & 32
& \bmstd{0.568}{0.280}
& \bmstd{0.232}{0.457}
& \bmstd{0.734}{0.355} \\

FBM & 64
& \mstd{0.488}{0.351}
& \mstd{0.168}{0.491}
& \mstd{0.755}{0.361} \\

FBM & 128
& \mstd{0.476}{0.321}
& \mstd{0.123}{0.437}
& \mstd{0.790}{0.339} \\

\midrule

TM & 12
& \mstd{0.474}{0.276}
& \mstd{0.118}{0.116}
& \mstd{0.411}{0.083} \\

TM & 16
& \bmstd{0.545}{0.225}
& \bmstd{0.194}{0.119}
& \bmstd{0.395}{0.078} \\

TM & 32
& \mstd{0.470}{0.306}
& \mstd{0.164}{0.090}
& \mstd{0.405}{0.080} \\

TM & 64
& \mstd{0.414}{0.372}
& \mstd{0.115}{0.146}
& \mstd{0.411}{0.085} \\

TM & 128
& \mstd{0.407}{0.361}
& \mstd{0.118}{0.091}
& \mstd{0.421}{0.084} \\

\midrule

FR & 12
& \bmstd{0.698}{0.096}
& \bmstd{0.272}{0.311}
& \bmstd{0.211}{0.027} \\

FR & 16
& \mstd{0.605}{0.173}
& \mstd{0.060}{0.545}
& \mstd{0.232}{0.044} \\

FR & 32
& \mstd{0.601}{0.151}
& \mstd{-0.009}{0.685}
& \mstd{0.239}{0.050} \\

FR & 64
& \mstd{0.576}{0.173}
& \mstd{-0.016}{0.703}
& \mstd{0.240}{0.055} \\

FR & 128
& \mstd{0.561}{0.170}
& \mstd{-0.047}{0.717}
& \mstd{0.246}{0.054} \\

\bottomrule
\end{tabular}

\end{table}

\begin{table}[t]
\centering
\caption{Model-capacity configurations used in the size ablation.
$L_{\mathrm{EEG}}$ and $L_{\mathrm{E2M}}$ denote the numbers of EEG-encoder
and total E2M Transformer layers, respectively. \rev{Parameters (M) use the full-model scope specified in Table~\ref{tab:sota_combined}.}}
\label{tab:model_size_config}

\small
\setlength{\tabcolsep}{4pt}
\begin{tabular}{llccccccc}
\toprule
Dataset & Size
& $d_{\mathrm{EEG}}$
& $d_{\mathrm{E2M}}$
& $L_{\mathrm{EEG}}$
& $L_{\mathrm{E2M}}$
& $H_{\mathrm{EEG}}$
& $H_{\mathrm{E2M}}$
& \rev{Params (M)} \\
\midrule

FBM & Large
& 512 & 512 & 6 & 16 & 8 & 8 & \rev{89.804} \\
& Middle
& 512 & 256 & 5 & 12 & 8 & 8 & \rev{45.384} \\
& Small
& 512 & 256 & 4 & 8 & 8 & 4 & \rev{39.073} \\
& Extra Small
& 512 & 256 & 3 & 6 & 8 & 4 & \rev{34.341} \\
\midrule

TM & Large
& 256 & 256 & 6 & 16 & 8 & 8 & \rev{22.458} \\
& Middle
& 256 & 256 & 5 & 12 & 8 & 8 & \rev{18.510} \\
& Small
& 256 & 128 & 4 & 8 & 8 & 4 & \rev{9.712} \\
& Extra Small
& 256 & 128 & 3 & 6 & 8 & 4 & \rev{8.526} \\
\midrule

FR & Large
& 256 & 256 & 6 & 16 & 8 & 8 & \rev{24.359} \\
& Middle
& 256 & 256 & 5 & 12 & 8 & 8 & \rev{20.410} \\
& Small
& 256 & 128 & 4 & 8 & 8 & 4 & \rev{11.653} \\
& Extra Small
& 256 & 128 & 3 & 6 & 8 & 4 & \rev{10.467} \\

\bottomrule
\end{tabular}
\end{table}

\begin{table*}[t]
\centering
\caption{Illustrative single-subject model-size ablation across three datasets. Results are reported on FBM (\texttt{S10}), TM (\texttt{S07}), and FR (\texttt{S05}).}
\label{tab:model_size_ablation}

\resizebox{\textwidth}{!}{%
\begin{tabular}{lccc|ccc|ccc}
\toprule
\multirow{2}{*}{Model Size}
& \multicolumn{3}{c|}{\textbf{FBM} (\texttt{S10})}
& \multicolumn{3}{c|}{\textbf{TM} (\texttt{S07})}
& \multicolumn{3}{c}{\textbf{FR} (\texttt{S05})} \\
\cmidrule(lr){2-4}
\cmidrule(lr){5-7}
\cmidrule(lr){8-10}

& PCC$\uparrow$ & $R^2\uparrow$ & RMSE$\downarrow$
& PCC$\uparrow$ & $R^2\uparrow$ & RMSE$\downarrow$
& PCC$\uparrow$ & $R^2\uparrow$ & RMSE$\downarrow$ \\
\midrule

Large
& \textbf{0.878} & \textbf{0.749} & \textbf{0.451}
& \textbf{0.803} & \textbf{0.323} & \textbf{0.305}
& \textbf{0.778} & \textbf{0.562} & \textbf{0.182} \\

Middle
& \underline{0.864} & 0.722 & \underline{0.467}
& 0.787 & \underline{0.288} & 0.312
& 0.769 & \underline{0.554} & \underline{0.184} \\

Small
& 0.861 & \underline{0.726} & 0.469
& \underline{0.797} & 0.282 & \underline{0.309}
& \underline{0.770} & 0.550 & 0.186 \\

Extra Small
& 0.856 & 0.714 & 0.469
& 0.791 & 0.283 & 0.311
& 0.766 & 0.545 & 0.187 \\

\bottomrule
\end{tabular}%
}
\end{table*}

\noindent\textbf{Effect of VQ codebook size.}
Table~\ref{tab:vq_codebook_ablation} expands capacity analysis to three subjects per dataset: FBM S02/S07/S10, TM S02/S06/S07, and FR S05/S07/S22. Under the same end-to-end evaluation, the default $K=32/16/12$ gives the best mean PCC, $R^2$, and RMSE for FBM/TM/FR. Increasing $K$ to 128 lowers PCC to 0.476/0.407/0.561. These results are consistent with a trade-off between motion expressiveness and EEG decodability. Smaller vocabularies may merge distinct motions, while finer vocabularies may create harder EEG discrimination tasks.

\noindent\textbf{Effect of model size.}
Tables~\ref{tab:model_size_config} and~\ref{tab:model_size_ablation}
compare four model-capacity settings. Starting from the default Large model,
we progressively reduce the EEG-encoder depth and E2M Transformer depth, and
for the smaller variants also reduce the E2M hidden width and number of
attention heads where applicable. The Large model achieves the best overall
performance across FBM, TM, and FR. Smaller models generally remain competitive,
although the ordering is not monotonic across all intermediate sizes. The
performance reduction is most visible on FBM, while TM and FR exhibit smaller
changes. Because these experiments use one representative subject per dataset,
they characterize sensitivity to model capacity rather than a population-level
scaling law.

\subsubsection{Temporal token granularity}
\label{app:token_granularity}
On the same three-subject subsets as the codebook study, we vary $N_m\in\{20,10,5\}$ and retrain all stages (Table~\ref{tab:token_granularity}). For 40-frame, 2-s windows, these settings yield nominal durations of 0.1/0.2/0.4\,s per token. The best mean results occur at 10 tokens for FBM/TM and 5 for FR, matching the defaults. Coarser tokenization can merge temporal phases; finer tokenization creates longer sequences and subtler adjacent distinctions. This sensitivity concerns the joint EEG-to-motion system.

\begin{table}[t]
\centering
\caption{Temporal token granularity on the same three-subject subsets as
Table~\ref{tab:vq_codebook_ablation}. All stages are retrained.
Each 2-s window contains 40 motion frames; $N_m=20/10/5$ corresponds to
$2\times/4\times/8\times$ compression. Entries are mean $\pm$
between-subject standard deviation.}
\label{tab:token_granularity}

\small
\setlength{\tabcolsep}{5pt}
\renewcommand{\arraystretch}{1.05}

\begin{tabular}{@{}llccc@{}}
\toprule
Dataset & $N_m$ & PCC$\uparrow$ & $R^2\uparrow$ & RMSE$\downarrow$ \\
\midrule

FBM & 20
& \mstd{0.512}{0.310}
& \mstd{0.195}{0.461}
& \mstd{0.747}{0.339} \\

FBM & 10
& \bmstd{0.568}{0.280}
& \bmstd{0.232}{0.457}
& \bmstd{0.734}{0.355} \\

FBM & 5
& \mstd{0.482}{0.373}
& \mstd{0.205}{0.463}
& \mstd{0.756}{0.371} \\

\midrule

TM & 20
& \mstd{0.503}{0.225}
& \mstd{0.155}{0.089}
& \mstd{0.405}{0.069} \\

TM & 10
& \bmstd{0.545}{0.225}
& \bmstd{0.194}{0.119}
& \bmstd{0.395}{0.078} \\

TM & 5
& \mstd{0.470}{0.259}
& \mstd{0.128}{0.106}
& \mstd{0.411}{0.077} \\

\midrule

FR & 20
& \mstd{0.489}{0.217}
& \mstd{0.005}{0.600}
& \mstd{0.249}{0.044} \\

FR & 10
& \mstd{0.511}{0.208}
& \mstd{0.040}{0.561}
& \mstd{0.240}{0.044} \\

FR & 5
& \bmstd{0.698}{0.096}
& \bmstd{0.272}{0.311}
& \bmstd{0.211}{0.027} \\

\bottomrule
\end{tabular}
\end{table}

\subsubsection{Alignment-weight sensitivity}
We jointly vary $\lambda_{\mathrm{pcl}}=\lambda_{\mathrm{supcon}}$ over $\{0.2,1,5\}$, with all remaining settings fixed and the same three-subject subsets. Table~\ref{tab:alignment_weights} favors weight 1 on all dataset means. Together with the individual loss removals in Table~\ref{tab:ablation_three_datasets}, this tests both component necessity and sensitivity to a common loss scale.
\begin{table}[t]
\centering
\caption{Joint alignment-weight sensitivity on the same three-subject
subsets as Table~\ref{tab:vq_codebook_ablation}, with
$\lambda_{\mathrm{pcl}}=\lambda_{\mathrm{supcon}}=\lambda$ and all other
settings fixed. Entries are mean $\pm$ between-subject standard deviation.}
\label{tab:alignment_weights}

\small
\setlength{\tabcolsep}{5pt}
\renewcommand{\arraystretch}{1.05}

\begin{tabular}{@{}llccc@{}}
\toprule
Dataset & $\lambda$ & PCC$\uparrow$ & $R^2\uparrow$ & RMSE$\downarrow$ \\
\midrule

FBM & 0.2
& \mstd{0.459}{0.407}
& \mstd{0.201}{0.472}
& \mstd{0.754}{0.363} \\

FBM & 1
& \bmstd{0.568}{0.280}
& \bmstd{0.232}{0.457}
& \bmstd{0.734}{0.355} \\

FBM & 5
& \mstd{0.443}{0.417}
& \mstd{0.201}{0.489}
& \mstd{0.762}{0.374} \\

\midrule

TM & 0.2
& \mstd{0.528}{0.232}
& \mstd{0.149}{0.117}
& \mstd{0.403}{0.077} \\

TM & 1
& \bmstd{0.545}{0.225}
& \bmstd{0.194}{0.119}
& \bmstd{0.395}{0.078} \\

TM & 5
& \mstd{0.472}{0.281}
& \mstd{0.126}{0.131}
& \mstd{0.410}{0.080} \\

\midrule

FR & 0.2
& \mstd{0.598}{0.159}
& \mstd{0.095}{0.507}
& \mstd{0.229}{0.045} \\

FR & 1
& \bmstd{0.698}{0.096}
& \bmstd{0.272}{0.311}
& \bmstd{0.211}{0.027} \\

FR & 5
& \mstd{0.618}{0.162}
& \mstd{0.154}{0.416}
& \mstd{0.225}{0.035} \\

\bottomrule
\end{tabular}
\end{table}

\subsubsection{Frozen versus trainable Stage-3 codebook}
\label{app:trainable_codebook}

We compare frozen and trainable Stage-3 codebooks on all 9/6/5 FBM/TM/FR subjects. The two variants share the same initialization, data, seed, motion-space MSE objective, optimizer, scheduler, iteration budget, and checkpoint criterion. Both update the EEG encoder, E2M Transformer, and VQ decoder while keeping the VQ encoder fixed; the trainable variant additionally updates the motion codebook. The frozen-codebook setting achieves higher mean PCC and $R^2$ and lower mean RMSE on all three datasets (Table~\ref{tab:trainable_codebook}), supporting the default choice of keeping the motion-prior interface fixed during Stage 3. 

\begin{table}[tbp]
\centering
\caption{Stage-3 codebook fine-tuning on the complete cohorts (FBM: 9; TM: 6; FR: 5 subjects). Entries are mean $\pm$ between-subject standard deviation. Both variants update the EEG encoder, E2M Transformer, and VQ decoder; only codebook trainability differs.}
\label{tab:trainable_codebook}

\small
\setlength{\tabcolsep}{5pt}
\renewcommand{\arraystretch}{1.05}

\begin{tabular}{@{}llccc@{}}
\toprule
Dataset & Codebook & PCC$\uparrow$ & $R^2\uparrow$ & RMSE$\downarrow$ \\
\midrule

FBM & Frozen
& \bmstd{0.522}{0.238}
& \bmstd{0.110}{0.354}
& \bmstd{0.718}{0.229} \\

FBM & Trainable
& \mstd{0.496}{0.270}
& \mstd{0.106}{0.362}
& \mstd{0.720}{0.230} \\

\midrule

TM & Frozen
& \bmstd{0.485}{0.211}
& \bmstd{0.184}{0.148}
& \bmstd{0.404}{0.052} \\

TM & Trainable
& \mstd{0.463}{0.222}
& \mstd{0.180}{0.147}
& \mstd{0.405}{0.051} \\

\midrule

FR & Frozen
& \bmstd{0.657}{0.152}
& \bmstd{0.250}{0.340}
& \bmstd{0.230}{0.038} \\

FR & Trainable
& \mstd{0.638}{0.153}
& \mstd{0.214}{0.341}
& \mstd{0.237}{0.039} \\

\bottomrule
\end{tabular}

\end{table}

\subsubsection{FBM refinement supervision scope}
\label{app:refinement_scope}
We compare default 30-D end-effector supervision with full-263-D supervision on all nine FBM subjects, retaining the same remaining settings. Table~\ref{tab:refinement_scope} shows that full supervision yields PCC/$R^2$/RMSE of $0.427\pm0.268/0.025\pm0.335/0.773\pm0.231$ and FID $4.844\pm5.494$, versus $0.522\pm0.238/0.110\pm0.354/0.718\pm0.229$ and FID $4.139\pm4.036$ for the default. Thus, full supervision recovers neither Stage-2 FID ($3.169\pm2.037$) nor default end-effector accuracy. Paired MSE and evaluator-space FID measure different properties, and the evaluator's domain mismatch further limits interpretation of small FID differences.
\begin{table}[tbp]
\centering
\caption{FBM Stage-3 supervision scope on all nine subjects. Both settings are evaluated with paired 30-D end-effector metrics and full-263-D FID. Entries are mean $\pm$ between-subject standard deviation; the remaining settings are identical.}
\label{tab:refinement_scope}

\small
\setlength{\tabcolsep}{5pt}
\renewcommand{\arraystretch}{1.05}

\begin{tabular}{@{}lcccc@{}}
\toprule
Supervised dimensions & PCC$\uparrow$ & $R^2\uparrow$ & RMSE$\downarrow$ & FID$\downarrow$ \\
\midrule

30-D (default)
& \mstd{0.522}{0.238}
& \mstd{0.110}{0.354}
& \mstd{0.718}{0.229}
& \mstd{4.14}{4.04} \\

263-D
& \mstd{0.427}{0.268}
& \mstd{0.025}{0.335}
& \mstd{0.773}{0.231}
& \mstd{4.84}{5.49} \\

\bottomrule
\end{tabular}

\end{table}

\subsubsection{\rev{Artifact-processed EEG Comparison}}
\label{app:artifact_processing}

\rev{We compare original and artifact-processed EEG on FBM S02/S07/S10, FR S05/S07/S22, and TM S01/S02/S05. TM uses EOG-reference adaptive denoising following \citet{he2018treadmill}; FR uses EOG-assisted ICA removal of ocular components; FBM uses the released processed EEG, including bad-channel handling, common-average referencing, and artifact subspace reconstruction~\citep{zhu2018fbm}. NeuroMotion, LR, and NeuroDyGait are retrained on the same processed EEG with the original session splits, motion targets, and evaluation metrics.}

\rev{Table~\ref{tab:artifact_processing} reports matched-subject results. NeuroMotion Stage 3 obtains processed-EEG mean PCC/$R^2$/RMSE of $0.494/0.181/0.761$ on FBM, $0.567/0.035/0.244$ on FR, and $0.199/0.010/0.452$ on TM, outperforming both comparison methods on all three cohort-mean reconstruction metrics. Its absolute reconstruction performance decreases relative to the matched original-EEG subsets, while its relative advantage is retained.}

\begin{table}[tbp]
\centering
\color{black}
\caption{Artifact-processing comparison on matched subjects: FBM S02/S07/S10, FR S05/S07/S22, and TM S01/S02/S05. Entries are mean $\pm$ between-subject standard deviation over three subjects. Original and processed settings use the same motion targets, splits, and evaluation metrics.}
\label{tab:artifact_processing}
\small
\setlength{\tabcolsep}{3pt}
\renewcommand{\arraystretch}{1.1}

\resizebox{\linewidth}{!}{%
\begin{tabular}{@{}llccc|ccc@{}}
\toprule
& & \multicolumn{3}{c|}{Original EEG}
& \multicolumn{3}{c}{Processed EEG} \\
Dataset & Method
& PCC$\uparrow$ & $R^2\uparrow$ & RMSE$\downarrow$
& PCC$\uparrow$ & $R^2\uparrow$ & RMSE$\downarrow$ \\
\midrule

FBM & NeuroMotion
& $\mathbf{0.568}_{\pm 0.280}$
& $\mathbf{0.232}_{\pm 0.457}$
& $\mathbf{0.734}_{\pm 0.355}$
& $\mathbf{0.494}_{\pm 0.328}$
& $\mathbf{0.181}_{\pm 0.389}$
& $\mathbf{0.761}_{\pm 0.274}$ \\

& NeuroDyGait
& $0.065_{\pm 0.042}$
& $-0.140_{\pm 0.112}$
& $0.995_{\pm 0.325}$
& $0.044_{\pm 0.032}$
& $-0.149_{\pm 0.103}$
& $0.998_{\pm 0.323}$ \\

& LR
& $0.295_{\pm 0.123}$
& $-1.203_{\pm 1.270}$
& $1.339_{\pm 0.717}$
& $0.215_{\pm 0.112}$
& $-0.914_{\pm 0.662}$
& $1.304_{\pm 0.500}$ \\

\midrule

FR & NeuroMotion
& $\mathbf{0.698}_{\pm 0.096}$
& $\mathbf{0.272}_{\pm 0.311}$
& $\mathbf{0.211}_{\pm 0.027}$
& $\mathbf{0.567}_{\pm 0.077}$
& $\mathbf{0.035}_{\pm 0.352}$
& $\mathbf{0.244}_{\pm 0.018}$ \\

& NeuroDyGait
& $0.633_{\pm 0.139}$
& $0.158_{\pm 0.398}$
& $0.223_{\pm 0.031}$
& $0.485_{\pm 0.113}$
& $-0.223_{\pm 0.722}$
& $0.259_{\pm 0.042}$ \\

& LR
& $0.391_{\pm 0.034}$
& $-0.185_{\pm 0.448}$
& $0.270_{\pm 0.020}$
& $0.216_{\pm 0.186}$
& $-1.392_{\pm 1.288}$
& $0.375_{\pm 0.088}$ \\

\midrule

TM & NeuroMotion
& $\mathbf{0.363}_{\pm 0.151}$
& $\mathbf{0.104}_{\pm 0.110}$
& $\mathbf{0.428}_{\pm 0.012}$
& $\mathbf{0.199}_{\pm 0.115}$
& $\mathbf{0.010}_{\pm 0.051}$
& $\mathbf{0.452}_{\pm 0.029}$ \\

& NeuroDyGait
& $0.327_{\pm 0.105}$
& $0.030_{\pm 0.049}$
& $0.444_{\pm 0.027}$
& $0.075_{\pm 0.109}$
& $-0.030_{\pm 0.023}$
& $0.461_{\pm 0.034}$ \\

& LR
& $0.111_{\pm 0.111}$
& $-1.473_{\pm 1.501}$
& $0.753_{\pm 0.223}$
& $0.041_{\pm 0.043}$
& $-6.243_{\pm 5.558}$
& $1.115_{\pm 0.399}$ \\

\bottomrule
\end{tabular}%
}
\end{table}

\subsection{Frequency-band Ablation Study}
\label{app:filtering}

We conduct a frequency-band ablation to examine which EEG spectral components are most informative for motion decoding under our experimental setting. For each subject and session, we apply band-pass filtering to the EEG signals and train/evaluate NeuroMotion using the same preprocessing, architecture, data split, and optimization protocol. We compare individual frequency bands, including \textit{delta}, \textit{theta}, \textit{alpha}, \textit{beta}, and \textit{gamma}, together with an \textit{all-band} setting using the full preprocessed EEG spectrum. Reconstruction performance is reported using PCC and RMSE, where higher PCC and lower RMSE indicate better decoding.

\begin{figure}[t]
  \centering
  \includegraphics[width=\linewidth]{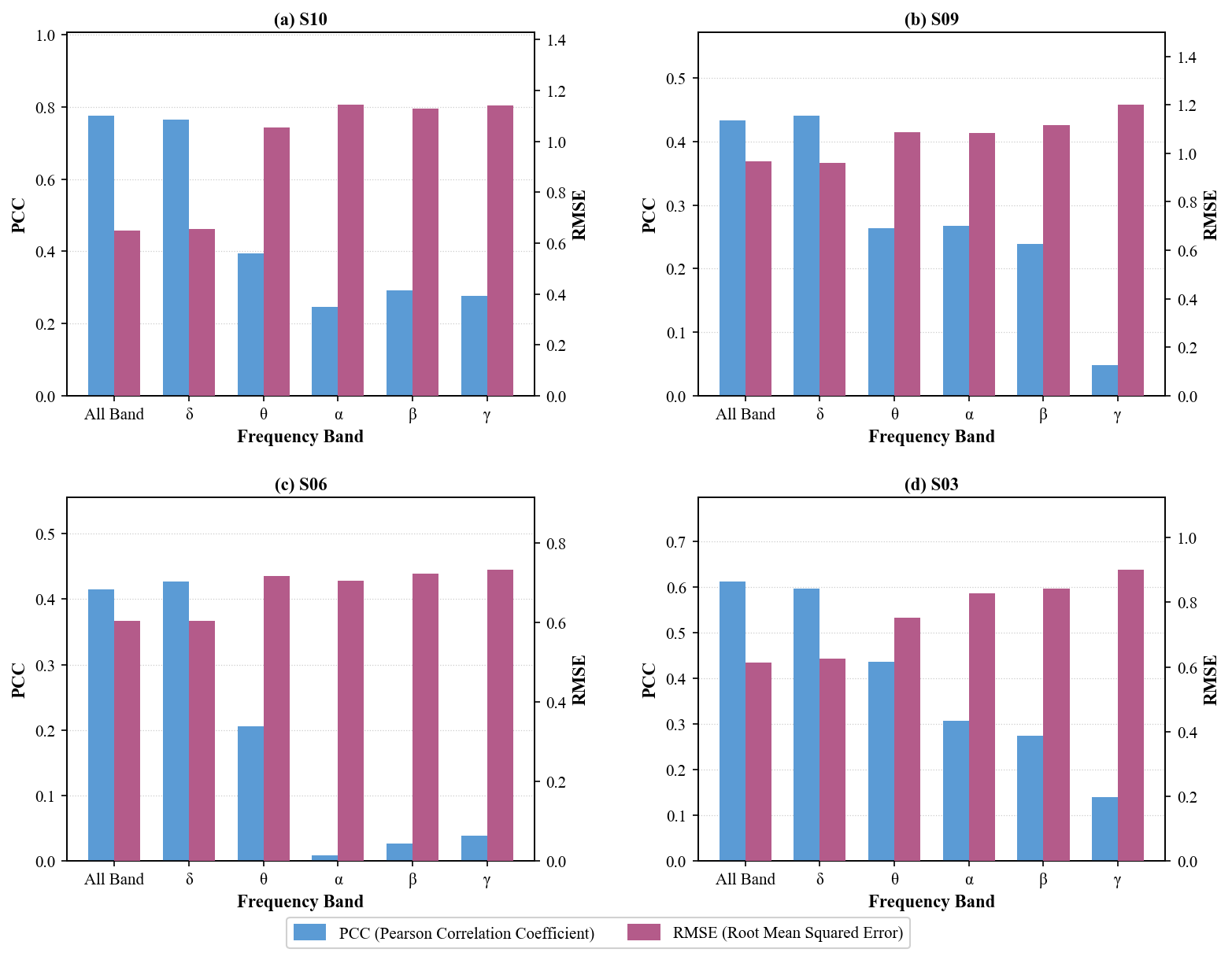}
  \caption{\textbf{Frequency-band analysis on FBM.} We report decoding performance under different EEG frequency-band settings, using PCC (higher is better) and RMSE (lower is better). Results for four FBM subjects (S10/S09/S06/S03) are summarized in a single panel, showing that low-frequency bands generally provide stronger decoding performance than higher-frequency bands.}
  \label{fig:app_freqband_fbm}
\end{figure}

Figure~\ref{fig:app_freqband_fbm} summarizes the FBM results. Low-frequency components provide the strongest predictive information under this setting. Delta-band and all-band EEG show higher mean PCC and lower mean RMSE than the higher-frequency bands, whereas alpha, beta, and gamma filtering generally degrades reconstruction performance. These results suggest that NeuroMotion relies more strongly on low-frequency EEG structure for naturalistic locomotion decoding, although the present analysis does not isolate neural activity from other low-frequency signal components.

This trend is consistent with prior EEG-based gait decoding studies. For example, ~\citet{luu2016unscented} showed that lower-limb joint angles during treadmill walking can be decoded from delta-band / slow-cortical-potential EEG using an Unscented Kalman filter. More broadly, low-frequency EEG has been associated with gait-cycle-related cortical dynamics and long-timescale sensorimotor integration during rhythmic movement~\citep{gwin2011electrocortical,knaepen2015grcp}. Higher-frequency components in mobile EEG can also be more affected by muscle activity and motion-related artifacts, which may reduce their reliability when used in isolation~\citep{delorme2004eeglab}.

\subsection{RSA Alignment Analysis}
\label{app:rsa_alignment}

\begin{figure*}[t]
  \centering
  \includegraphics[width=\textwidth]{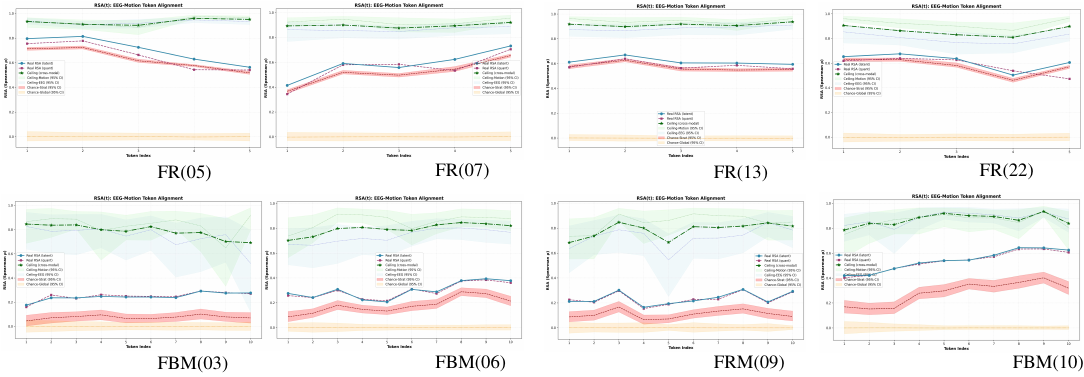}
  \caption{Token-level RSA alignment analysis on FBM and FR. Each curve shows the RSA score over the motion-token timeline, computed between EEG representations and motion-token representations. We visualize results for four FBM subjects (\texttt{S03}, \texttt{S06}, \texttt{S09}, \texttt{S10}) and four FR subjects (\texttt{sub-05}, \texttt{sub-07}, \texttt{sub-13}, \texttt{sub-22}).}
  \label{fig:app_rsa_token}
\end{figure*}

We further analyze neuro-motion alignment using representational similarity analysis (RSA)~\citep{kriegeskorte2008rsa}. We compare the representational structure of EEG embeddings and motion representations by computing representational dissimilarity matrices (RDMs) in the two spaces and reporting their Spearman correlation at each token position. Figure~\ref{fig:app_rsa_token} reports token-level RSA curves on FBM subjects \texttt{S03}, \texttt{S06}, \texttt{S09}, and \texttt{S10}, and FR subjects \texttt{S05}, \texttt{S07}, \texttt{S13}, and \texttt{S22}.

In the RSA plots, \textit{Real RSA (latent)} measures the correspondence between EEG embeddings and continuous motion latents before quantization, while \textit{Real RSA (quant)} measures the correspondence with quantized codebook embeddings. \textit{Chance-Global} globally permutes the correspondence between EEG and motion representations, whereas \textit{Chance-Strat} permutes EEG--motion correspondences within motion-code clusters, providing a structure-preserving null baseline. The \textit{Ceiling} curve provides an empirical reliability ceiling estimated from split-half consistency. Across subjects, the real RSA curves are generally above the chance baselines, supporting structural correspondence between the learned EEG representations and the motion-primitive space.

\begin{figure*}[t]
  \centering
  \includegraphics[width=\textwidth]{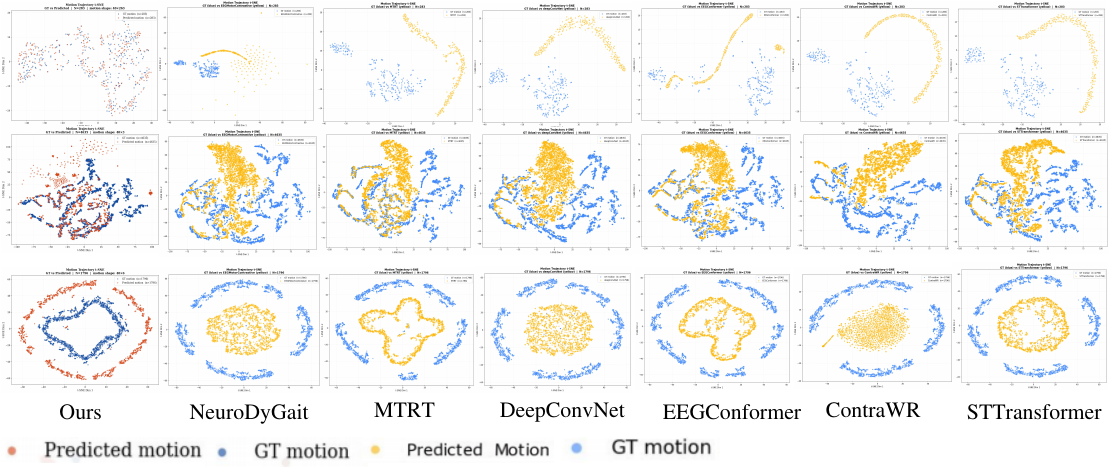}
  \caption{Trajectory distribution visualization using t-SNE on three representative subjects: FBM \texttt{S10}, FR \texttt{S05}, and TM \texttt{S07}. For each dataset, we compare the distributions of ground-truth trajectories, NeuroMotion predictions, and baseline predictions in a shared two-dimensional embedding space. NeuroMotion generally shows stronger overlap with the ground-truth distribution than the baselines.}
  \label{fig:app_traj_tsne}
\end{figure*}

\subsection{\rev{Decoded Motion-Code Visualization}}
\label{app:codebook_visualization}

\begin{figure}[tbp]
    \centering
    \IfFileExists{fbm_codebook_decodings.pdf}{%
        \includegraphics[width=0.96\linewidth]{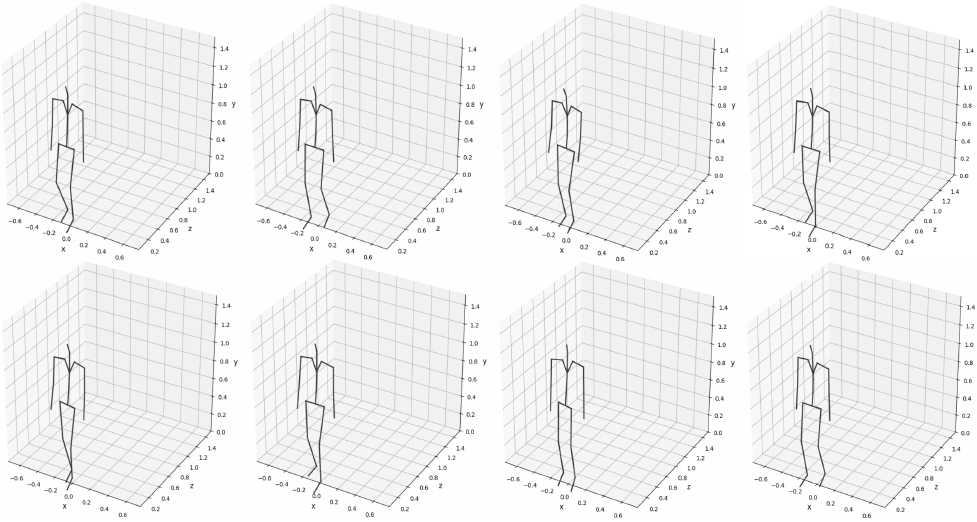}%
    }{%
        {\color{black}\fbox{%
            \parbox[c][42mm][c]{0.92\linewidth}{%
                \centering
                Eight representative FBM motion-code decodings\\[4pt]
                Placeholder for a $2\times4$ panel figure\\[4pt]
                Second decoded frame of each selected code
            }%
        }}%
    }
    \caption{\rev{Decoded poses of eight representative FBM motion-code entries. Each panel shows the second frame obtained by decoding one code in isolation.}}
    \label{fig:fbm_codebook_decodings}
\end{figure}

\rev{Figure~\ref{fig:fbm_codebook_decodings} visualizes the decoded outputs of eight representative FBM motion codes. Each selected code is supplied individually to the motion decoder, and its second decoded frame is shown as a representative pose. During trajectory reconstruction, the decoder processes a complete token sequence through temporal convolution and upsampling, so the local output associated with a code also depends on its neighboring codes.}

\subsection{Trajectory Distribution Visualization}
\label{app:traj_tsne}

We further visualize the distribution of generated and ground-truth trajectories using t-SNE. For each dataset, predicted and real trajectory windows from the held-out test session are flattened into vectors and jointly embedded into a two-dimensional space. This visualization is intended as a qualitative diagnostic of distributional overlap rather than a quantitative metric.

Figure~\ref{fig:app_traj_tsne} compares NeuroMotion with the baselines on three representative test subjects: FBM \texttt{S10}, FR \texttt{S05}, and TM \texttt{S07}. Across datasets, baseline predictions often form compact clusters or shifted regions relative to the ground-truth trajectories, indicating limited coverage of the real motion distribution. In contrast, NeuroMotion shows stronger overlap with the ground-truth distribution, suggesting that it better preserves the distributional structure of the target motion space.


\subsection{Full Subject-wise Baseline Comparisons}
\label{app:subjectwise_baselines}

To complement the cohort-level results, Tables~\ref{tab:subjectwise_fbm},
\ref{tab:subjectwise_fbm_dist}, \ref{tab:subjectwise_fr}, and
\ref{tab:subjectwise_tm} report complete subject-wise comparisons.
\rev{The added LR, RR, and UKF results use the same subjects, splits, targets, and metrics as Table~\ref{tab:sota_combined}. The pre-existing paired tests in Appendix~\ref{app:cohort_statistics} retain their original deep-learning-baseline comparison set.}
Reconstruction entries are reported as PCC/$R^2$/RMSE.
Bold and underlined values denote the best and second-best results, respectively,
determined using the original unrounded values.
For FBM distributional metrics, FID is ranked in the same manner, whereas
Diversity is reported without ranking because it should be interpreted relative
to the real-motion reference.

\begin{table}[t]
\centering
\caption{
Full subject-level reconstruction results on FBM.
PCC, $R^2$, and RMSE are reported in three vertically stacked panels.
Bold and underlined values denote the best and second-best results,
respectively, determined using the original unrounded values.
}
\label{tab:subjectwise_fbm}

\small
\renewcommand{\arraystretch}{1.05}
\setlength{\tabcolsep}{2pt}

\begin{tabular*}{\linewidth}{
@{\extracolsep{\fill}}lccccccccc@{}
}
\multicolumn{10}{c}{\textbf{(a) PCC $\uparrow$}} \\[2pt]
\toprule
Method & s02 & s03 & s04 & s05 & s06 & s07 & s08 & s09 & s10 \\
\midrule
\rev{LR} & \rev{$\underline{0.344}$} & \rev{$\underline{0.411}$} & \rev{$\underline{0.136}$} & \rev{$-0.084$} & \rev{$0.186$} & \rev{$0.155$} & \rev{$0.126$} & \rev{$\underline{0.303}$} & \rev{$0.385$} \\
\rev{RR} & \rev{$\mathbf{0.345}$} & \rev{$\underline{0.411}$} & \rev{$\underline{0.136}$} & \rev{$-0.084$} & \rev{$0.187$} & \rev{$\underline{0.157}$} & \rev{$0.126$} & \rev{$\underline{0.303}$} & \rev{$\underline{0.386}$} \\
\rev{UKF} & \rev{$0.188$} & \rev{$0.392$} & \rev{$0.082$} & \rev{$-0.097$} & \rev{$\underline{0.257}$} & \rev{$0.066$} & \rev{$0.129$} & \rev{$0.267$} & \rev{$0.339$} \\
DCN & $-0.002$ & $0.010$ & $0.022$ & $0.014$ & $0.006$ & $0.032$ & $\underline{0.135}$ & $0.037$ & $0.037$ \\
EC & $0.035$ & $0.023$ & \rev{$0.027$} & $0.017$ & $0.012$ & $0.009$ & $0.039$ & $0.024$ & $0.050$ \\
ContraWR & $0.018$ & $0.002$ & $0.006$ & $-0.012$ & $0.026$ & \rev{$0.066$} & $0.074$ & $0.004$ & $0.054$ \\
DY & \rev{$0.091$} & \rev{$0.068$} & \rev{$0.027$} & $0.029$ & \rev{$0.046$} & $0.017$ & $0.054$ & $0.022$ & \rev{$0.088$} \\
ST & $0.029$ & $0.023$ & $0.022$ & $0.009$ & $0.006$ & $0.020$ & $0.020$ & $0.009$ & $0.010$ \\
MTRT & $0.039$ & $0.055$ & $0.023$ & $\underline{0.032}$ & $0.014$ & $0.035$ & $0.022$ & \rev{$0.041$} & $0.034$ \\
\midrule
Ours & \rev{$0.333$} & $\mathbf{0.725}$ & $\mathbf{0.345}$ & $\mathbf{0.093}$ & $\mathbf{0.674}$ & $\mathbf{0.494}$ & $\mathbf{0.526}$ & $\mathbf{0.632}$ & $\mathbf{0.878}$ \\
\bottomrule
\end{tabular*}

\vspace{8pt}

\begin{tabular*}{\linewidth}{
@{\extracolsep{\fill}}lccccccccc@{}
}
\multicolumn{10}{c}{\textbf{(b) $R^2$ $\uparrow$}} \\[2pt]
\toprule
Method & s02 & s03 & s04 & s05 & s06 & s07 & s08 & s09 & s10 \\
\midrule
\rev{LR} & \rev{$-0.588$} & \rev{$-0.226$} & \rev{$-1.446$} & \rev{$-2.238$} & \rev{$-1.060$} & \rev{$-2.664$} & \rev{$-3.130$} & \rev{$-0.654$} & \rev{$-0.358$} \\
\rev{RR} & \rev{$-0.583$} & \rev{$-0.224$} & \rev{$-1.434$} & \rev{$-2.230$} & \rev{$-1.054$} & \rev{$-2.590$} & \rev{$-3.114$} & \rev{$-0.650$} & \rev{$-0.352$} \\
\rev{UKF} & \rev{$-0.128$} & \rev{$\underline{0.027}$} & \rev{$-0.399$} & \rev{$-0.705$} & \rev{$-0.337$} & \rev{$-3.330$} & \rev{$-0.576$} & \rev{$\underline{-0.102}$} & \rev{$\underline{0.036}$} \\
DCN & $-0.203$ & $-0.114$ & $\underline{-0.147}$ & $\underline{-0.091}$ & $-0.381$ & $-0.292$ & $-0.444$ & $-0.515$ & $-0.166$ \\
EC & $-0.156$ & $-0.220$ & $-0.447$ & $-0.252$ & $-0.493$ & $-0.384$ & $-0.422$ & \rev{$-0.123$} & $-0.263$ \\
ContraWR & $-0.153$ & $-0.092$ & $-0.323$ & $-0.165$ & $-0.372$ & $-0.282$ & $\underline{-0.395}$ & $-0.489$ & $-0.083$ \\
DY & $-0.043$ & $-0.039$ & $\mathbf{-0.107}$ & $-0.093$ & $\underline{-0.251}$ & $\underline{-0.263}$ & $-0.684$ & $-0.313$ & $-0.115$ \\
ST & $-0.138$ & $-0.091$ & $-0.338$ & $-0.205$ & $-0.493$ & $-0.369$ & $-0.498$ & $-0.893$ & $-0.259$ \\
MTRT & $\underline{-0.016}$ & \rev{$-0.015$} & $-0.148$ & $\mathbf{-0.088}$ & $-0.311$ & $\underline{-0.263}$ & $-0.569$ & $-0.546$ & \rev{$-0.048$} \\

\midrule
Ours & $\mathbf{0.068}$ & $\mathbf{0.474}$ & $-0.330$ & $-0.104$ & $\mathbf{0.143}$ & $\mathbf{-0.120}$ & $\mathbf{-0.228}$ & $\mathbf{0.337}$ & $\mathbf{0.749}$ \\
\bottomrule
\end{tabular*}

\vspace{8pt}

\begin{tabular*}{\linewidth}{
@{\extracolsep{\fill}}lccccccccc@{}
}
\multicolumn{10}{c}{\textbf{(c) RMSE $\downarrow$}} \\[2pt]
\toprule
Method & s02 & s03 & s04 & s05 & s06 & s07 & s08 & s09 & s10 \\
\midrule
\rev{LR} & \rev{$0.755$} & \rev{$0.816$} & \rev{$1.102$} & \rev{$1.200$} & \rev{$0.779$} & \rev{$2.140$} & \rev{$1.781$} & \rev{$1.240$} & \rev{$1.123$} \\
\rev{RR} & \rev{$0.753$} & \rev{$0.815$} & \rev{$1.099$} & \rev{$1.201$} & \rev{$0.776$} & \rev{$2.120$} & \rev{$1.770$} & \rev{$1.245$} & \rev{$1.121$} \\
\rev{UKF} & \rev{$0.674$} & \rev{$\underline{0.721}$} & \rev{$0.818$} & \rev{$0.865$} & \rev{$\underline{0.616}$} & \rev{$2.480$} & \rev{$1.190$} & \rev{$\underline{1.040}$} & \rev{$\underline{0.968}$} \\
DCN & $1.063$ & $0.793$ & $\underline{0.762}$ & $\underline{0.697}$ & $0.646$ & $1.313$ & $\underline{1.124}$ & $1.109$ & $1.051$ \\
EC & $0.682$ & $0.821$ & $0.837$ & $0.741$ & $0.673$ & $1.340$ & $1.137$ & $1.050$ & $1.092$ \\
ContraWR & $0.683$ & $0.785$ & $0.816$ & $0.719$ & $0.637$ & $\underline{1.300}$ & $1.134$ & \rev{$1.045$} & $1.017$ \\
DY & $0.653$ & $0.770$ & $\mathbf{0.750}$ & $0.698$ & \rev{$0.618$} & $1.301$ & $1.171$ & $1.078$ & $1.030$ \\
ST & $0.678$ & $0.785$ & $0.812$ & $0.729$ & $0.668$ & $1.337$ & $1.155$ & $1.217$ & $1.090$ \\
MTRT & $\underline{0.647}$ & \rev{$0.763$} & $0.764$ & $\mathbf{0.697}$ & $0.629$ & $1.310$ & $1.155$ & $1.111$ & \rev{$1.005$} \\

\midrule
Ours & $\mathbf{0.619}$ & $\mathbf{0.536}$ & $0.829$ & $0.702$ & $\mathbf{0.465}$ & $\mathbf{1.133}$ & $\mathbf{0.949}$ & $\mathbf{0.783}$ & $\mathbf{0.451}$ \\
\bottomrule
\end{tabular*}
\end{table}

\begin{table}[t]
\centering
\caption{Full subject-level distributional results on FBM. Each entry reports FID/Diversity. Lower FID is better; Diversity is reported without ranking because it is interpreted relative to the real-motion reference.}
\label{tab:subjectwise_fbm_dist}
\scriptsize
\setlength{\tabcolsep}{1.5pt}
\renewcommand{\arraystretch}{1.05}

\resizebox{\linewidth}{!}{%
\begin{tabular}{@{}lccccccccc@{}}
\toprule
Method & s02 & s03 & s04 & s05 & s06 & s07 & s08 & s09 & s10 \\
\midrule
\rev{LR} & \rev{$4.540/3.204$} & \rev{$2.170/3.635$} & \rev{$56.660/8.060$} & \rev{$3.930/3.163$} & \rev{$16.600/8.080$} & \rev{$10.740/7.160$} & \rev{$2.039/3.110$} & \rev{$41.700/8.529$} & \rev{$\underline{11.940}/8.140$} \\
\rev{RR} & \rev{$4.510/3.190$} & \rev{$2.150/3.630$} & \rev{$56.500/8.050$} & \rev{$3.925/3.160$} & \rev{$\underline{16.500}/8.070$} & \rev{$\underline{10.600}/7.170$} & \rev{$\underline{2.020}/3.100$} & \rev{$41.700/8.520$} & \rev{$\underline{11.940}/8.145$} \\
\rev{UKF} & \rev{$\mathbf{0.889}/1.913$} & \rev{$\underline{1.894}/2.520$} & \rev{$\mathbf{10.040}/4.240$} & \rev{$\underline{0.991}/2.100$} & \rev{$\underline{16.500}/7.710$} & \rev{$19.260/7.740$} & \rev{$2.682/2.177$} & \rev{$59.100/7.230$} & \rev{$21.500/7.600$} \\

DCN & $73.8/0.339$ & $45.2/2.79$ & $37.8/3.61$ & $48.9/3.03$ & $91.3/3.77$ & $57.0/5.30$ & $10.6/1.07$ & $70.7/1.92$ & $128/4.46$ \\
EC & \rev{$2.66/0.647$} & \rev{$5.84/0.429$} & \rev{$22.4/2.84$} & \rev{$1.91/0.665$} & \rev{$23.8/2.51$} & $30.9/1.84$ & $9.60/0.389$ & $25.1/2.84$ & $77.8/7.06$ \\
ContraWR & $44.0/3.50$ & $57.0/3.02$ & $22.6/0.849$ & $52.3/3.53$ & $71.3/2.65$ & $55.4/5.58$ & $27.5/3.28$ & $61.5/0.235$ & $120/2.69$ \\
DY & $46.8/2.57$ & $50.8/2.40$ & $38.5/5.14$ & $41.7/2.47$ & $88.0/4.95$ & $74.8/5.59$ & $19.9/3.01$ & $76.9/4.01$ & $149/5.28$ \\
ST & $2.86/0.635$ & \rev{$5.84/0.429$} & $77.0/4.68$ & $3.03/0.665$ & $36.6/3.67$ & $62.7/4.76$ & \rev{$9.48/0.602$} & $67.7/0.263$ & \rev{$77.7/7.01$} \\
MTRT & $23.4/1.57$ & $21.2/1.52$ & $30.7/4.95$ & $25.3/1.90$ & $36.9/5.16$ & \rev{$22.9/5.75$} & $13.3/1.59$ & $\underline{21.7}/3.36$ & $104/5.03$ \\

\midrule
Ours & \rev{$\underline{1.24}/1.74$} & $\mathbf{1.70}/2.22$ & \rev{$\underline{12.0}/4.66$} & $\mathbf{0.867}/1.33$ & $\mathbf{2.86}/7.69$ & $\mathbf{5.05}/6.58$ & $\mathbf{1.90}/2.15$ & $\mathbf{1.92}/3.57$ & $\mathbf{9.67}/9.33$ \\
\bottomrule
\end{tabular}%
}
\end{table}

\begin{table}[t]
\centering
\caption{Full subject-level reconstruction results on FR. Each entry reports PCC/$R^2$/RMSE. Bold and underlined values denote the best and second-best results, respectively, determined using the original unrounded values.}
\label{tab:subjectwise_fr}
\scriptsize
\setlength{\tabcolsep}{2pt}
\renewcommand{\arraystretch}{1.08}

\resizebox{\linewidth}{!}{%
\begin{tabular}{@{}lccccc@{}}
\toprule
Method & s05 & s07 & s13 & s15 & s22 \\
\midrule
\rev{LR} & \rev{$0.428/0.044/0.274$} & \rev{$0.385/-0.702/0.288$} & \rev{$0.302/-0.763/0.476$} & \rev{$0.016/-1.736/0.430$} & \rev{$0.361/0.102/0.249$} \\
\rev{RR} & \rev{$0.416/0.006/0.279$} & \rev{$0.335/-0.793/0.297$} & \rev{$0.335/-0.793/0.297$} & \rev{$0.018/-1.857/0.439$} & \rev{$0.319/0.046/0.257$} \\
\rev{UKF} & \rev{$0.241/-1.021/0.400$} & \rev{$0.256/-1.000/0.315$} & \rev{$0.237/-2.110/0.633$} & \rev{$-0.040/-8.578/0.812$} & \rev{$0.272/-0.359/0.306$} \\

DCN & $0.708/0.438/0.205$ & $0.614/-0.583/0.274$ & $0.658/0.374/0.284$ & $0.146/-0.434/0.311$ & $0.406/0.132/0.244$ \\
EC & $0.727/0.441/0.206$ & $0.493/-0.852/0.298$ & $0.635/0.302/0.300$ & $0.120/-0.652/0.333$ & $0.397/-0.009/0.262$ \\
ContraWR & $0.683/0.259/0.239$ & $\underline{0.712}/-0.103/\mathbf{0.234}$ & $0.646/0.351/0.289$ & $-0.022/-0.807/0.342$ & $\underline{0.548}/\underline{0.270}/\underline{0.225}$ \\
DY & $\underline{0.740}/\underline{0.527}/\underline{0.189}$ & $0.683/-0.264/0.248$ & $\underline{0.749}/\underline{0.507}/\underline{0.252}$ & $\underline{0.390}/\underline{-0.174}/\mathbf{0.278}$ & $0.476/0.212/0.232$ \\
ST & $0.714/0.425/0.209$ & $0.490/-0.468/0.276$ & $0.628/0.312/0.298$ & $0.152/-0.790/0.340$ & $0.459/0.190/0.237$ \\
MTRT & $0.721/0.463/0.203$ & $0.552/\mathbf{-0.036}/0.237$ & $0.626/0.366/0.286$ & $0.093/-0.449/0.313$ & $0.533/0.210/0.232$ \\

\midrule
Ours & $\mathbf{0.778}/\mathbf{0.562}/\mathbf{0.182}$ & $\mathbf{0.724}/\underline{-0.057}/\underline{0.235}$ & $\mathbf{0.773}/\mathbf{0.581}/\mathbf{0.232}$ & $\mathbf{0.421}/\mathbf{-0.146}/\underline{0.287}$ & $\mathbf{0.592}/\mathbf{0.312}/\mathbf{0.217}$ \\
\bottomrule
\end{tabular}%
}
\end{table}

\begin{table}[t]
\centering
\caption{Full subject-level reconstruction results on TM. Each entry reports PCC/$R^2$/RMSE. Bold and underlined values denote the best and second-best results, respectively, determined using the original unrounded values.}
\label{tab:subjectwise_tm}
\scriptsize
\setlength{\tabcolsep}{2pt}
\renewcommand{\arraystretch}{1.08}

\resizebox{\linewidth}{!}{%
\begin{tabular}{@{}lcccccc@{}}
\toprule
Method & s01 & s02 & s05 & s06 & s07 & s08 \\
\midrule
\rev{LR} & \rev{$0.231/-0.149/0.504$} & \rev{$0.091/-1.167/0.934$} & \rev{$0.012/-3.104/0.820$} & \rev{$0.201/-0.305/0.531$} & \rev{$0.183/-5.730/1.063$} & \rev{$0.377/0.136/0.463$} \\
\rev{RR} & \rev{$0.196/-0.327/0.541$} & \rev{$0.018/-4.844/0.981$} & \rev{$0.019/-4.844/0.981$} & \rev{$0.148/-0.623/0.591$} & \rev{$0.307/0.043/0.486$} & \rev{$0.307/0.044/0.487$} \\
\rev{UKF} & \rev{$0.167/-0.337/0.541$} & \rev{$0.055/-6.520/1.310$} & \rev{$0.001/-6.171/1.033$} & \rev{$0.196/-0.419/0.547$} & \rev{$0.307/0.044/0.486$} & \rev{$0.370/0.127/0.466$} \\
DCN & $0.384/\underline{0.084}/\underline{0.449}$ & $0.126/-0.237/0.535$ & $0.111/-0.295/0.466$ & $\mathbf{0.447}/\mathbf{0.165}/\mathbf{0.423}$ & $0.633/0.111/0.372$ & $\mathbf{0.632}/0.314/0.413$ \\
EC & $0.005/-0.051/0.484$ & $0.340/-0.084/0.500$ & $0.088/-0.047/0.420$ & $0.419/0.122/0.434$ & $0.598/0.004/0.389$ & $0.548/0.251/0.431$ \\
ContraWR & $-0.030/-0.004/0.479$ & $0.004/-0.028/0.487$ & $0.012/-0.045/0.420$ & $-0.003/-0.028/0.471$ & $-0.017/-0.237/0.464$ & $0.000/-0.006/0.501$ \\
DY & $\underline{0.391}/0.029/0.459$ & $\underline{0.385}/\underline{0.080}/\underline{0.460}$ & $\underline{0.206}/\mathbf{-0.018}/\underline{0.412}$ & $\underline{0.447}/\underline{0.164}/\underline{0.423}$ & $0.653/\underline{0.142}/\underline{0.359}$ & $0.617/\underline{0.363}/\underline{0.398}$ \\
ST & $0.072/-0.032/0.479$ & $0.300/-0.057/0.494$ & $\mathbf{0.218}/\underline{-0.019}/\mathbf{0.411}$ & $0.427/0.139/0.430$ & $\underline{0.703}/0.089/0.364$ & $0.545/0.260/0.427$ \\
MTRT & $0.000/-0.024/0.478$ & $0.001/-0.010/0.483$ & $0.005/-0.037/0.419$ & $0.026/-0.046/0.475$ & $0.007/-0.267/0.467$ & $0.384/0.038/0.489$ \\

\midrule
Ours & $\mathbf{0.456}/\mathbf{0.167}/\mathbf{0.430}$ & $\mathbf{0.444}/\mathbf{0.169}/\mathbf{0.438}$ & $0.189/-0.023/0.415$ & $0.388/0.089/0.443$ & $\mathbf{0.803}/\mathbf{0.323}/\mathbf{0.305}$ & $\underline{0.632}/\mathbf{0.379}/\mathbf{0.392}$ \\
\bottomrule
\end{tabular}%
}
\end{table}

\begin{table}[!ht]
\centering
\caption{Default-model size and inference resources on one NVIDIA L4 (batch 1, FP32). MACs are per EEG window; memory is peak allocated GPU memory.}
\label{tab:inference_resources}

\vspace{-1pt}

\small
\setlength{\tabcolsep}{5pt}
\renewcommand{\arraystretch}{1.0}

\begin{tabular}{@{}lcccc@{}}
\toprule
Dataset & Params (M) & CKPT (MiB) & MACs (G) & Peak (MiB) \\
\midrule
FBM & $80.217$ & $306.17$ & $1.430$ & $325.4$ \\
TM  & $20.156$ & $77.05$  & $0.373$ & $150.1$ \\
FR  & $21.008$ & $80.29$  & $0.214$ & $126.5$ \\
\bottomrule
\end{tabular}

\vspace{-6pt}
\end{table}

\begin{table}[!ht]
\centering
\caption{Offline inference timing on one NVIDIA L4: 100 warm-ups and five repeats of 1,000 measurements. Model latency is GPU forward time; post-window latency includes CPU--GPU transfers from preprocessed CPU EEG to reconstructed CPU motion.}
\label{tab:inference_latency}

\vspace{-1pt}

\small
\setlength{\tabcolsep}{5pt}
\renewcommand{\arraystretch}{1.0}

\begin{tabular}{@{}lccc@{}}
\toprule
Dataset & Model mean/P95 (ms) & Post-window mean/P95 (ms) & Windows/s \\
\midrule
FBM & $17.750/20.071$ & $18.082/21.610$ & $56.84$ \\
TM  & $17.474/18.286$ & $17.572/18.420$ & $57.91$ \\
FR  & $17.102/17.770$ & $17.289/18.143$ & $57.80$ \\
\bottomrule
\end{tabular}

\vspace{-6pt}
\end{table}

\subsection{Cohort Statistics and Paired Comparisons}
\label{app:cohort_statistics}

Tables~\ref{tab:subjectwise_fbm}, \ref{tab:subjectwise_fr}, and
\ref{tab:subjectwise_tm} provide the complete subject-level results for all
compared methods. Subjects, rather than windows or output dimensions, are
treated as the statistical unit.For the paired comparisons reported below, 
we compare NeuroMotion Stage 3
with NeuroDyGait, the strongest overall baseline among the original
deep-learning methods, using two-sided paired $t$-tests across subjects.

On FBM, NeuroMotion outperforms the overall strongest baseline, NeuroDyGait, on 9/9
subjects for PCC and 7/9 subjects for both $R^2$ and RMSE. The corresponding
$p$-values are 0.00027, 0.0229, and 0.0246, respectively. On TM, NeuroMotion
outperforms NeuroDyGait on 4/6 subjects for each metric, with $p=0.291$,
$0.202$, and $0.222$ for PCC, $R^2$, and RMSE, respectively. On FR, NeuroMotion
outperforms NeuroDyGait on 5/5 subjects for PCC and $R^2$ and 4/5 subjects
for RMSE, with corresponding $p$-values of 0.0407, 0.0519, and 0.125.
The reported $p$-values are uncorrected for multiple comparisons, and the
small TM and FR cohorts limit statistical power.

Relative to Stage 2, Stage 3 improves all three reconstruction metrics for
9/9 FBM and 6/6 TM subjects. On FR, PCC improves for 5/5 subjects and
$R^2$/RMSE for 4/5, with \texttt{S07} unchanged on the latter two metrics.

\subsection{Inference Cost and Timing Scope}
\label{app:inference_cost}

We profile the default models offline on a single NVIDIA L4 GPU using batch-1 FP32 PyTorch eager execution, with 100 warm-up runs and five repeats of 1,000 measurements. Tables~\ref{tab:inference_resources} and~\ref{tab:inference_latency} report the number of parameters, serialized checkpoint size, MACs per input window, peak allocated GPU memory, mean/P95 latency, and throughput.\rev{The full-model parameter counts in Tables~\ref{tab:sota_combined}
and~\ref{tab:model_size_config} include the frozen motion encoder,
whereas the inference-model counts in Table~\ref{tab:inference_resources}
exclude it}. Model latency measures GPU forward execution only. Post-window latency starts from a preprocessed EEG window in CPU memory and ends with the reconstructed motion returned to CPU memory, including host--device transfers.

Across the three default models, the largest post-window P95 latency is 21.610\,ms on FBM, with peak allocated GPU memory of 325.4\,MiB. These measurements characterize only the model-side computation after a preprocessed 2-s EEG window becomes available; they exclude EEG acquisition, preprocessing, feedback, and control. No complete online closed-loop BCI experiment has been conducted. End-to-end acquisition-to-action latency, feedback stability, and online calibration therefore remain to be evaluated.

\clearpage
\section{Model Architecture Details}
\label{app:model_details}

\subsection{EEG Encoder Architecture}
\label{app:eeg_encoder_arch}

\noindent\textbf{Input/Output.}
The EEG encoder maps a raw EEG window $\mathbf{X}\in\mathbb{R}^{B\times C\times T}$ to a sequence of EEG tokens
$\mathbf{E}\in\mathbb{R}^{B\times N_e\times d_e}$, where the number of EEG channels $C$ and hidden dimension $d_e$ are dataset-dependent.
Unless otherwise stated, we use dropout $p{=}0.5$ throughout the encoder.

\begin{table*}[t]
\centering
\caption{EEG encoder architecture. $C$ denotes the number of EEG channels, $N_e$ denotes the number of EEG tokens, and $d_e$ denotes the EEG embedding dimension. These values are dataset-dependent.}
\label{tab:app_eeg_encoder_arch}
\setlength{\tabcolsep}{5.9pt}
\renewcommand{\arraystretch}{1.05}
\begin{tabular}{p{0.34\textwidth} p{0.60\textwidth}}
\toprule
\textbf{Component} & \textbf{Architecture} \\
\midrule

Input reshape &
(0): Unsqueeze channel dimension \\
& \hspace{1.0em}$[B,C,T]\rightarrow[B,1,C,T]$ \\[2pt]

Temporal patch embedding &
(1): Conv2d$(1,40,\ \text{kernel}=(1,10),\ \text{stride}=(1,1),\ \text{pad}=(0,0))$ \\
& \hspace{1.0em}$[B,1,C,T]\rightarrow[B,40,C,T-9]$ \\[2pt]

Channel fusion &
(2): Conv2d$(40,40,\ \text{kernel}=(C,1),\ \text{stride}=(1,1),\ \text{pad}=(0,0))$ \\
& \hspace{1.0em}$[B,40,C,T-9]\rightarrow[B,40,1,T-9]$ \\[2pt]

Normalization + activation &
(3): BatchNorm2d$(40)$ $\rightarrow$ ELU \\
& \hspace{1.0em}$[B,40,1,T-9]\rightarrow[B,40,1,T-9]$ \\[2pt]

Adaptive pooling &
(4): AdaptiveAvgPool2d$((1,N_e))$ \\
& \hspace{1.0em}$[B,40,1,T-9]\rightarrow[B,40,1,N_e]$ \\[2pt]

Patch dropout &
(5): Dropout$(p{=}0.5)$ \\[2pt]

Projection to model dimension &
(6): Conv2d$(40,d_e,\ \text{kernel}=(1,1),\ \text{stride}=(1,1))$ \\
& \hspace{1.0em}$[B,40,1,N_e]\rightarrow[B,d_e,1,N_e]$ \\[2pt]

Rearrange to token sequence &
(7): Rearrange ``$b\ e\ 1\ w \rightarrow b\ w\ e$'' \\
& \hspace{1.0em}$[B,d_e,1,N_e]\rightarrow[B,N_e,d_e]$ \\[2pt]

Learned positional embedding &
(8): $\text{Pos}\in\mathbb{R}^{1\times N_e\times d_e}$, added to EEG tokens \\
& \hspace{1.0em}$[B,N_e,d_e]\rightarrow[B,N_e,d_e]$ \\[2pt]

Transformer encoder blocks $\times 6$ &
(9--14): Repeat for $l{=}1\ldots 6$: \\
& \hspace{1.0em}LayerNorm$(d_e)$ \\
& \hspace{1.0em}MultiHeadAttention$(\text{dim}=d_e,\ \text{heads}=8,\ \text{attn\_drop}=0.5)$ \\
& \hspace{1.0em}Residual + Dropout$(p{=}0.5)$ \\
& \hspace{1.0em}LayerNorm$(d_e)$ \\
& \hspace{1.0em}FFN: Linear$(d_e,4d_e)$ $\rightarrow$ GELU $\rightarrow$ Dropout$(0.5)$ $\rightarrow$ Linear$(4d_e,d_e)$ \\
& \hspace{1.0em}Residual \\

\bottomrule
\end{tabular}
\end{table*}

\subsection{E2M Transformer Architecture}
\label{app:e2m_transformer_arch}

\noindent\textbf{Input/Output.}
The E2M module performs parallel motion-token prediction conditioned on EEG tokens.
Given EEG features $\mathbf{E}\in\mathbb{R}^{B\times N_e\times d_e}$, we first project them to the E2M hidden dimension $d_m$.
We then concatenate the projected EEG tokens with a set of learnable target slots
$\mathbf{Q}\in\mathbb{R}^{N_m\times d_m}$, expanded along the batch dimension.
The resulting sequence length is $L=N_e+N_m$.
A single Transformer forward pass outputs logits over the motion codebook, and the predicted motion tokens are obtained from the final $N_m$ target-slot positions. The additive attention mask on the concatenated sequence is
\[
M_{ij}=\begin{cases}0,&j\leq i,\\-\infty,&j>i.\end{cases}
\]
Thus each target slot attends to the full EEG prefix and preceding/current slots. Predictions are not fed back, and all target logits are computed in one pass. The EEG encoder itself uses unmasked attention. Learned EEG positional embeddings fit fixed-length observations; fixed sinusoidal E2M positions order conditioning and target slots without additional learned parameters. This positional choice is pragmatic, not a separate contribution.

\begin{table}[tbp]
\centering
\caption{Tensor notation. Mathematical code IDs are one-based. We configure $N_e=N_m=N$ and $d_e=d_c$, while keeping the symbols distinct to identify their roles.}
\label{tab:tensor_notation}
\small
\begin{tabular}{lll}
\toprule
Symbol & Meaning & Shape/domain \\
\midrule
$H$ & EEG encoder output & $B\times N_e\times d_e$ \\
$H^{\mathrm{e2m}}$ & Adapted EEG tokens & $B\times N_e\times d_m$ \\
$Q^B$ & Batched target slots & $B\times N_m\times d_m$ \\
$Z^{\mathrm{e2m}}$ & E2M hidden states & $B\times(N_e+N_m)\times d_m$ \\
$P$ & Motion-token logits & $B\times N_m\times K$ \\
$\mathcal{C}$ & Motion codebook & $K\times d_c$ \\
$s_{b,i}$ & Discrete code index & $\{1,\ldots,K\}$ \\
$G$ & One-hot selections & $B\times N_m\times K$ \\
\bottomrule
\end{tabular}
\end{table}

\begin{table*}[t]
\centering
\caption{E2M Transformer architecture used for parallel motion-token decoding. $N_e$ denotes the number of EEG tokens, $N_m$ denotes the number of motion-token slots, $d_e$ denotes the EEG encoder output dimension, $d_m$ denotes the E2M hidden dimension, and $K$ denotes the motion codebook size. In our implementation, $N_m$ is matched to $N_e$.}
\label{tab:app_e2m_transformer_arch_parallel}
\setlength{\tabcolsep}{3.9pt}
\renewcommand{\arraystretch}{1.05}
\begin{tabular}{p{0.24\textwidth} p{0.72\textwidth}}
\toprule
\textbf{Component} & \textbf{Architecture} \\
\midrule

EEG conditioning projection &
(0): Linear$(d_e,d_m)$ \\
& \hspace{1.2em}Input: $\mathbf{E}\in\mathbb{R}^{B\times N_e\times d_e}$, Output: $\mathbb{R}^{B\times N_e\times d_m}$ \\[2pt]

Learnable target slots &
(1): Unified learnable token embeddings $\mathbf{Q}\in\mathbb{R}^{N_m\times d_m}$ \\
& \hspace{1.2em}The target slots are used across all subjects and expanded to $\mathbb{R}^{B\times N_m\times d_m}$. \\[2pt]

Concatenation &
(2): Concatenate projected EEG tokens and target slots along the token dimension: \\
& \hspace{1.2em}$\mathbb{R}^{B\times N_e\times d_m}\oplus \mathbb{R}^{B\times N_m\times d_m}
\rightarrow \mathbb{R}^{B\times (N_e+N_m)\times d_m}$ \\[2pt]

Sine-cosine position encoding &
(3): Fixed sin-cos positional encoding for length $L=N_e+N_m$, added to the concatenated sequence. \\[2pt]

Dropout &
(4): Dropout$(p{=}0.5)$ \\[2pt]

Causal self-attention blocks $\times N_{\mathrm{e2m}}$ &
(5): Repeat for $l=1,\ldots,N_{\mathrm{e2m}}$: \\
& \hspace{1.2em}LayerNorm$(d_m)$ \\
& \hspace{1.2em}CausalSelfAttention$(\text{dim}=d_m,\ \text{heads}=h,\ \text{attn\_drop}=0.5)$ with causal mask \\
& \hspace{1.2em}Residual + Dropout$(p{=}0.5)$ \\
& \hspace{1.2em}LayerNorm$(d_m)$ \\
& \hspace{1.2em}MLP: Linear$(d_m,4d_m)$ $\rightarrow$ GELU $\rightarrow$ Linear$(4d_m,d_m)$ $\rightarrow$ Dropout$(0.5)$ \\
& \hspace{1.2em}Residual \\[2pt]

Final normalization &
(7): LayerNorm$(d_m)$ \\[2pt]

Vocabulary head &
(8): Linear$(d_m,K)$ (bias=False), where $K$ is the motion codebook size \\
& \hspace{1.2em}Output logits: $\mathbb{R}^{B\times (N_e+N_m)\times K}$; the final $N_m$ positions correspond to predicted motion tokens. \\

\bottomrule
\end{tabular}
\end{table*}

\clearpage
\section{Further Discussion}
\subsection{Limitations.}
First, our experiments are restricted to a small number of subjects, reflecting the current scarcity of synchronized EEG--motion datasets and limiting population-level generalization~\citep{zhu2018fbm,mohanta2025move4as}.
Second, our motion interface relies on a VQ-VAE tokenizer: discretizing continuous human motion into a finite codebook provides a strong inductive bias and stabilizes long-horizon generation, but it also introduces quantization error and restricts expressivity to the learned vocabulary, potentially attenuating fine-grained kinematics and stylistic nuances. 
Encouragingly, recent text-to-motion models have begun to reflect on this limitation and explore alternatives that interface autoregressive backbones with \emph{continuous} motion latents (e.g., VAE latents) and use lightweight diffusion modules for high-fidelity synthesis, mitigating the symbolic--continuous mismatch induced by tokenization~\citep{zhu2025motiongpt3}. 
Adapting such continuous-latent or diffusion-bridged motion priors to EEG-conditioned generation is a promising direction, but requires revisiting token-level alignment objectives and efficient online inference.
Finally, TM/FR cross-day evaluation and offline L4 profiling address limited aspects of generalization and efficiency. FBM has no verified day metadata, long-term drift is untested, and a bidirectional encoder requires the available observation window. Streaming acquisition, feedback stability, and online user adaptation remain open challenges.

\subsection{Broader implications.} 
Broadly, this work suggests that successful neural decoding of complex behavior may require representational alignment rather than direct signal-to-trajectory mapping~\citep{farshchian2022nomad}.
Our framework offers a general blueprint for linking neural data to structured generative models, potentially extending beyond motion to other high-dimensional behavioral or perceptual domains.



\end{document}